\documentclass{article} % For LaTeX2e
\PassOptionsToPackage{numbers,sort&compress}{natbib} % NHUT
\usepackage{iclr2025_conference,times}

\setcitestyle{numbers,square}
\makeatletter
\let\citet\citep
\makeatother

\usepackage{amsmath,amsfonts,bm}

\def\eqref#1{equation~\ref{#1}}
\def\1{\bm{1}}

\DeclareMathAlphabet{\mathsfit}{\encodingdefault}{\sfdefault}{m}{sl}
\SetMathAlphabet{\mathsfit}{bold}{\encodingdefault}{\sfdefault}{bx}{n}

\usepackage{hyperref}
\usepackage{url}
\usepackage{graphicx}
\usepackage{multirow}
\usepackage{fancyhdr}
\usepackage{booktabs}
\usepackage{subcaption}
\usepackage{algorithm}
\usepackage{algorithmic}
\usepackage{gensymb}
\usepackage{amsmath}
\usepackage{setspace}
\usepackage{amssymb}
\usepackage{tikz}
\usepackage[percent]{overpic}
\usepackage{comment}
\usepackage{float}
\usepackage{microtype}
\usepackage[normalem]{ulem}
\usepackage{enumitem}

\title{Beyond topography: Topographic regularization improves robustness and reshapes representations in convolutional neural networks}

\author{Nhut Truong \& Uri Hasson \\
Center for Mind/Brain Sciences (CIMeC), University of Trento\\
Rovereto, 38068, Italy \\
\texttt{leminhnhut.truong@unitn.it}, \quad \texttt{uri.hasson@unitn.it}
}

\iclrfinalcopy % Uncomment for camera-ready version, but NOT for submission.
\begin{document}

\maketitle

\vspace{0.95em}

\begin{abstract}
Topographic convolutional neural networks (TCNNs) are computational models that can simulate aspects of the brain’s spatial and functional organization. However, it is unclear whether and how different types of topographic regularization shape robustness, representational structure, and functional organization during end-to-end training. We address this question by comparing TCNNs trained with two local spatial losses applied to a penultimate-layer topographic grid: \textit{i}) Weight Similarity (WS), whose objective penalizes differences between neighboring units’ incoming weight vectors, and \textit{ii}) Activation Similarity (AS), whose objective penalizes differences between neighboring units’ activation patterns over stimuli. We evaluate the trained models on classification accuracy, robustness to weight perturbations and input degradation, the spatial organization of learned representations, and development of category-selective ``expert units'' in the penultimate layer. Both losses changed inter-unit correlation structure, but in qualitatively different ways. WS produced smooth topographies, with correlated neighborhoods. In contrast, AS produced a bimodal inter-unit correlation structure that lacked spatial smoothness. AS and WS training increased robustness relative to control (non-topographic) models: AS improved robustness to image degradation on CIFAR-10, WS did so on MNIST, and both improved robustness to weight perturbations. WS was also associated with greater input sensitivity at the unit level and stronger functional localization. In addition, as compared to control models, both AS and WS produced differences in orientation tuning, symmetry sensitivity, and eccentricity profiles of units. Together, these results show that local topographic regularization can improve robustness during end-to-end training while systematically reshaping representational structure.
\end{abstract}

\noindent Code is available on Github:  \url{github.com/tlmnhut/weight_vs_act_toponet}

\newpage
\section{Introduction}

Topographic models are computational models that provide an analogy to the cortical organization of the brain by instantiating a cortical-like map. In these models, each unit is assigned a position in a 2D grid \citep{poli2023introducing}, which defines a notion of distance between grid units. A topographic loss term can then be defined, which encourages spatially local similarity or smoothness in units’ responses (or parameters) across the grid. It has been shown that when a task loss (e.g., cross-entropy) is optimized jointly with a topographic loss, both shallow \citep{jacobs1992computational} and deep \citep{blauch2022connectivity, margalit2024unifying, lu2025end, qian2024local, doshi2023, deb2025toponets, zhang2024biologically, rathi2024topolm, binhuraib2025topoformer, al2025end} topographic networks can produce competitive task performance, while producing spatially organized responses that resemble cortical functional maps. For example, they produce angular and orientation preference such as those observed in early visual cortex, as well as category-selective clusters analogous to those reported in higher-level visual cortex \citep[e.g.,][]{blauch2022connectivity, margalit2024unifying, lu2025end, qian2024local, doshi2023, deb2025toponets, zhang2024biologically}. These models can also capture spatial biases in human behavior, recognizing objects more accurately when they appear in locations where they are most frequently experienced \citep{lu2025end}. Outside the vision domain, topographic modeling has reproduced tonotopic organization in auditory cortex \citep{al2025end}, and higher-level language organization \cite[e.g.,][]{rathi2024topolm, binhuraib2025topoformer}. Topographic networks can also be more strongly pruned, suggesting a sparser distribution of weight values \citep{poli2023introducing, deb2025toponets}.

The correlated activation patterns produced by topographic models carry parallels to brain activity, where nearby neurons often fire together, with correlations sometimes exceeding $r=0.4$ in certain cortical circuits \citep[e.g.,][]{zohary1994correlated, hansen2012correlated}.  Although such correlations reduce the degrees of freedom and representational capacity in biological neural populations \citep{zohary1994correlated}, they may also offer benefits such as robustness, as redundant neurons can compensate for one another \citep{harris2013cortical}. A similar trade-off has been discussed for computational models: on the one hand correlated units encode overlapping information and can impair decoding performance \citep{abbott1999effect}, but on the other, moderate correlations can act as a form of regularization, producing more compact representations or giving priority to more informative features \citep{poli2023introducing}. In summary, correlations constrain capacity but may provide computational advantages.

While topographic models are a topic of emerging interest in computational and systems neuroscience \citep{margalit2024unifying, rathi2024topolm, lu2025end, lee2020, blauch2022connectivity, doshi2023, zhang2024biologically, krug2023visualizing, hannagan2021, keller2021, qian2024local, jiang2024, deb2025toponets, dehghani2024credit, finzi2023single, achterberg2023spatially, binhuraib2025topoformer, zhou2025tdsnns, bashivan2025learning, Jozwik2023FirstSI, mehrer2020individual, al2025end, cortinovis2025investigating}, there is a substantial gap in our understanding of the computational consequences of imposing topographic regularization. This is an important open question not only for computational neuroscience, where topographic networks are mainly studied for their ability to reproduce brain-like spatial organization of responses, but also for machine learning, where the impacts of these locally induced correlations are poorly understood. 

Going beyond a focus on spatial organization, we therefore ask whether these correlations provide any computational advantages, such as improved robustness to noise, and how they, in turn, shape the representations learned by topographic networks. To our knowledge, neither of these issues has been systematically examined in controlled comparisons to date. Specifically, we address these questions using two related criteria:

\begin{enumerate}[leftmargin=*] 
\item \textbf{Network robustness}: We introduce internal noise by perturbing the readout weight matrix  connecting the penultimate (topographic) layer to the classification layer \citep{cheney2017robustness, arechiga2018effect, savva2023robustness}, and external noise by corrupting input images at test. We test how resilient topographic models are compared to non-topographic models in terms of both representational stability and classification accuracy
. 
\item \textbf{Representational properties}: We compare topographic models to size-matched non-topographic control models; we quantify unit-level sparsity and entropy, similarity of weights and activations, the dimensionality of the latent space, and the tendency of models to produce category-selective ``expert units''. We further characterize the spatial organization of representations by measuring the smoothness of topographic maps and the extent to which activity is functionally localized (i.e., whether similarly responding units are spatially clustered). Finally, we test whether topographic regularization changes functional organization by quantifying orientation, eccentricity, and angular tuning in topographic and non-topographic networks.
\end{enumerate}

To understand the computational effects of topographic regularization, we use local spatial losses and end-to-end training. Prior work has often relied on global spatial losses that include a distance-dependent term over all unit pairs. These global losses directly encourage high activation similarity between nearby units as well as low activation similarity between distant ones \citep[e.g.,][]{poli2023introducing, margalit2024unifying}. They therefore effectively \textit{impose} functional localization by suppressing long-range correlations. This makes localized clusters an outcome of the training objective rather than an emergent property. We therefore restrict the regularizer to local interactions only \citet[e,g.,][]{lu2025end}. In addition, we train topographic networks end-to-end, because post hoc projection methods operating on frozen pretrained features \citep[e.g., self-organized mappings,][]{doshi2023, jiang2024, krug2023visualizing, zhang2024biologically} cannot address whether topographic regularization impacts feature learning itself.

\begin{figure}[]
    \includegraphics[width=\linewidth]{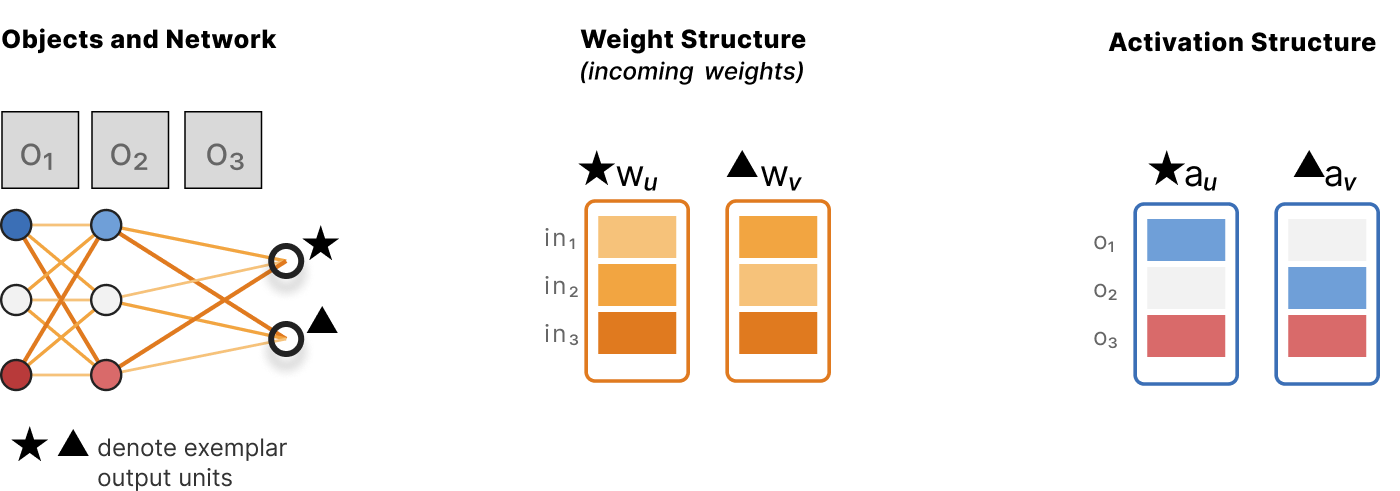}
    \caption{\textbf{Overview of main concepts}. Objects ($O_1$--$O_3$) are passed to a neural network. Two example output units ($\star = u$ and $\blacktriangle = v$) are indicated. For each example unit, an \emph{activation vector} ($\mathbf{a}_u, \mathbf{a}_v$) summarizes that unit’s responses across objects, and a \emph{weight vector} ($\mathbf{w}_u, \mathbf{w}_v$) summarizes the incoming weights from the preceding layer. Activation similarity is computed by correlating activation vectors across objects. Weight similarity is computed as a distance between weight vectors.}
    \label{fig:overview}
\end{figure}

Regarding the technical implementation, correlated activations in topographic models are often produced using an objective that encourages similarity between neighboring units’ activation patterns \citep[e.g.,][]{lee2020, margalit2024unifying, rathi2024topolm, poli2023introducing}. An alternative is to impose similarity constraints on incoming weights \citep{lu2025end}, which may be more biologically plausible: correlated activity should emerge due to shared afferent connectivity instead of being directly enforced at the level of activations. For this reason we also avoid explicit lateral connections, as correlations in the brain often reflect shared afferent input \citep{shadlen1998variable}. To directly contrast these two positions we compare two local topographic losses (Figure~\ref{fig:overview}): \textbf{Activation Similarity (AS)}, whose objective is to produce correlated activations in adjacent units, and \textbf{Weight Similarity (WS)}, whose objective is to produce similar afferent weight vectors in adjacent units. In both cases, we consider only local neighbors.

We find that AS and WS training produces qualitatively different computational outcomes. Compared to AS and control (non-topographic) models, WS training produces more representations that are more robust to perturbations of the readout weight matrix or degraded inputs. WS also produced greater input sensitivity, seen in higher activation entropy and lower activation sparsity, and stronger functional localization, with similarly responding units positioned at closer spatial distances. 

As compared to control models, AS and WS produced different emphases on angular, orientation and symmetry tuning. Interestingly this was not necessarily accompanied by a reduced effective dimensionality of the network activation, as for the lowest level of the topographic regularizer, the overall effective dimensionality of topographic models matched or exceeded those of the control models. Importantly, while WS training produced spatially smooth activations on a 2D grid, AS training did not, instead producing a striped-like spatial organization of activation. Taken together, this suggests that topographic regularization does not simply reorganize the weight matrix to maintain classification, but produces a markedly different feature space. In addition, our findings indicate that topographic regularization does not necessarily produce spatially smooth activation maps, as the networks can satisfy the objective via alternative solutions. Thus, not all topographic objectives are necessarily relevant models for cortical organization. 

% CAN REMOVE; SHOUL DIn sum, by conducting representational analyses commonly used in topographic modelling and robustness evaluations, our study not only systematically characterizes topography under different spatial constraints, but also highlights its potential engineering benefits. In doing so, we aim to bring this research line in computational cognitive neuroscience to the broader machine learning community.

\section{Related work}
\subsection{Topographic regularization in artificial neural networks (ANNs)}
Several studies have examined how spatially organized response patterns, often accompanied by increased local correlations, can be induced in ANNs by applying topographic objectives. One approach is end-to-end topographic training, which introduces a topographic loss term directly into the training objective. Typically, a joint loss combines a task loss (e.g., cross-entropy for classification) with an additional topographic spatial term.  Early work \citet{jacobs1992computational} introduced a spatial loss in small-scale, fully connected networks, penalizing the weight magnitude proportionally to the physical distances between the pre- and post-synaptic units’ assigned positions. As a result, adjacent units showed similar tuning. Others \citet[e.g.,][]{poli2023introducing} introduced a loss that enforces an inverse relationship between the activation similarity of spatially arranged convolutional filters and their distances. This improved robustness to pruning while maintaining task performance. Related approaches that produce cortical-like spatial organization include penalizing connections between spatially-remote units \citep{blauch2022connectivity}, introducing lateral pooling between neighboring filters \citep{qian2024local}, and adding topographic structure to variational autoencoders \citep{keller2021}.
Topography can also emerge without an explicit spatial loss when kernels in deeper layers are defined as averages of partially overlapping kernels from a preceding layer, producing smooth transitions across kernel maps \citep{bashivan2025learning}.

Other work used a hybrid approach \citet{margalit2024unifying}, in which unit locations  are optimized based on activation similarity computed from a pretrained model and then held these locations fixed during subsequent training. This model reproduced spatial properties of both early and high-level visual cortical features and outperformed standard networks on biological benchmarks. Similarly, \citet{lu2025end} trained a fully topographic model by maximizing weight vector similarity between immediate neighboring units, and showed that the resulting networks capture spatial biases including center-periphery preferences.

Another class of topographic models is based on post hoc projection of pretrained representations, for example, via self-organizing maps (SOMs). In these approaches, a fixed, pretrained feature space is mapped onto a 2D grid with the objective of producing spatial clusters of similarly-responding units \citep{doshi2023, jiang2024, krug2023visualizing, zhang2024biologically}. Conceptually, these methods learn a spatial embedding of \textit{existing} representations: they operate on frozen features and therefore cannot address how topographic regularization shapes feature learning during end-to-end task training \citep{aflalo2006}. 
% For this reason, they are not the focus of the current study.

\subsection{Impact of topography on representational structure}
Beyond producing correlated responses, topographic regularization also influences the representational structure of ANNs. Several studies have reported that such regularization reduces the effective dimensionality of latent representations \citep{deb2025toponets, margalit2024unifying, qian2024local} and improves robustness to pruning \citep{poli2023introducing, deb2025toponets}. 

At the same time, inter-unit correlations are sometimes considered detrimental in machine learning research. For example, the Barlow Twins architecture \citep{zbontar2021barlow} maximizes agreement between paired views (original and distorted image) while penalizing correlated activity across units as computed from the off-diagonal terms of the batch-wise cross-correlation matrix. The study demonstrates that reducing correlations improves classification accuracy. In addition similarity between incoming weight vectors is often considered a negative computational property, as minimizing such correlations improves classification accuracy \citep{cogswell2015reducing, rodriguez2016regularizing, jin2020does, wang2020mma}. Reducing redundancy across convolutional filters has been reported to produce similar benefits \citep{zhang2025compact}.

\section{Methods}
\subsection{Models and datasets}
\paragraph{MNIST.} 
The model used for MNIST \citep{lecun1998mnist} training  was a relatively shallow CNN. It consisted of two convolutional layers: the first with 32 output channels and the second with 64 channels, each using a \(3 \times 3\) kernel. Both convolutional layers were followed by a ReLU activation function and \(2 \times 2\) max-pooling. After the second convolutional layer (\(\texttt{conv2}\)), global average pooling was applied to each of the 64 feature maps, producing a 64-dimensional feature vector for each input image. This vector was then fed into a fully connected layer, \(\texttt{fc1}\), which mapped the 64-dimensional vector to 121 units. The output of \(\texttt{fc1}\) was connected to a second fully connected layer, \(\texttt{fc2}\), which produced the final 10 logits for classification, corresponding to the 10 MNIST classes. To reduce overfitting, dropout with a rate of 0.5 was applied after \(\texttt{fc1}\).  
% The control model was identical to the topographic models with the only exception being that the 121 \texttt{fc1} units not treated as a grid ad a spatial loss function was not applied

\paragraph{CIFAR-10.}
We used the standard CIFAR-10, a 10-class dataset of $32\times32$ images \citep{krizhevsky2009learning}. The CNN model used for image classification consisted of four convolutional layers with batch normalization applied after each. The number of channels in the first three layers were: 32, 64, and 128, each followed by max-pooling $stride=2$. The fourth convolutional layer consisted of 256 channels and was followed by a global average pooling layer ($n=256$ values). The last two layers were two fully connected layers. The first (\texttt{fc1}) consisted of 121 units, with dropout (0.3), and the second (\texttt{fc2}) mapped feature activations to the 10 output classes. 

The exact same model definitions were used for the topographic models and the control models, for CIFAR-10 and MNIST. The only difference was that in the topographic models, the 121 \texttt{fc1} units were arranged on a $11\times11$ grid to which a spatial loss function could be applied as described below.

\subsection{Spatial loss: weight-similarity and activation-similarity}
\paragraph{Weight similarity.} For training with a weight-similarity objectives, we used a joint loss function, combining the standard cross-entropy loss term, \(\mathcal{L}_{\mathrm{CE}} = \text{cross-entropy}(\text{output}, \text{target})\), and a spatial loss term \(\mathcal{L}_{\mathrm{spatial}}\). For weight-similarity, the spatial loss term was designed to increase similarity among immediately adjacent weight vectors in the \(11 \times 11\) grid structure.  To compute \(\mathcal{L}_{\text{spatial}}\), we indexed the 121 incoming weight vectors of \texttt{fc1} on an \(11 \times 11\) grid. Each grid position corresponds to one \texttt{fc1} unit with an incoming weight vector of dimension 64 (MNIST) or 256 (CIFAR-10). For each of the 121 grid cells, the immediate neighbors were identified. Then, for each cell, the \( L_2 \) norm (Euclidean distance) was computed between the weight vector of that cell and those of each neighboring cell. These distances were summed across all cells and divided by the number of neighboring cells, producing a single value indicating the average pairwise distance across the grid. The joint loss function was therefore \(\mathcal{L}_{\text{joint}} = \mathcal{L}_{\text{CE}} + \lambda\, \mathcal{L}_{\text{spatial}}\), where \(\lambda\) is a weighting factor that sets the contribution of the spatial loss. We evaluated the impact of the spatial loss term under six weighting levels, with $\lambda \in \{0.1, 0.3, 0.5, 1, 2, 3\}$.

\paragraph{Activation similarity.} The activation-similarity (AS) spatial loss term produced a single value indicating the similarity of each unit to its immediate neighbors. For AS, similarity was defined as  $1 - r$, where \( r \) is the Pearson's correlation between the two units’ activation vectors across minibatch samples. Here too \texttt{fc1} was the topographic layer, defined by reshaping 121 units to \(11 \times 11\) grid.

\paragraph{Training parameters.} MNIST models were trained using the Adam optimizer with a learning rate of $\eta = 0.001$ for 15 epochs. Initial evaluations showed that all models converged to a training-set accuracy of approximately $97\%$ under moderate regularization strength. The CIFAR-10 model was trained using the same optimizer parameters for 30 epochs, reaching similar training accuracy of around 96\% across the three model conditions. 

We trained 10 independently initialized models for each $\lambda$ level under both WS and AS regularization, resulting in 60 WS models and 60 AS models in total. Additionally, we trained 10 control models without topographic regularization, which are equivalent to $\lambda = 0$. In all analyses, mean statistics were computed from each set of 10 models from the same $\lambda$.

\subsection{Robustness tests}

\paragraph{Robustness of representational geometry.} After training the control, AS and WS models on MNIST and CIFAR-10, we extracted the weight matrix connecting the topographic penultimate layer to the classification layer (a $10 \times 121$ matrix in both cases). We consider each of this  matrix's 10 row vectors as a \textit{class weight vector; (CWVs)}. From these CWVs, we computed a class-weight representational similarity matrix (RSM), which is a $10 \times 10$ matrix of pairwise similarity between CWVs \citep{lake2015deep, nayak2019zero, filus2024extracting, filus2025doggest}. This was treated as the baseline representational geometry. 

% corresponds to a category and is often interpreted as a class prototype \citep{lake2015deep, nayak2019zero, filus2024extracting, filus2025doggest}. From these prototype vectors, we computed a representational similarity matrix (RSM), which is a $10 \times 10$ matrix of pairwise similarity between class prototypes. This was treated as the baseline representational geometry. 

To evaluate the robustness of this representation, we conducted several perturbation tests in which Gaussian noise (four intensity levels) was added to the $10 \times 121$ class-weight matrix, and the RSM was recomputed. We refer to these as perturbed RSMs. Each analysis was repeated 100 times, and we report the mean results.

We determined the impact of noise using two metrics. First, we computed the second-order similarity as the cosine similarity between the upper triangles of the unperturbed and perturbed RSMs. This indicates how much the representational geometry was impacted by noise. This was computed separately for the WS, AS, and control models. Second, we evaluated the drop in classification accuracy caused by the addition of noise, relative to the unperturbed models.

\paragraph{Robustness to input corruption.} For both WS and AS models, we evaluated their performance under various corruptions, with the noise applied to test-set images (training images were not corrupted). In all cases, corrupted images were normalized to the mean and standard deviation of MNIST and CIFAR-10 training sets. The noise interventions consisted of adding white noise, pink noise, and salt-and-pepper noise. White noise was introduced by adding to each pixel a random value from the standard normal distribution. Pink (1/f) noise was generated such that the power spectral density of the signal is inversely proportional to its frequency. Salt-and-pepper noise converted a proportion of randomly chosen image pixels to either black or white. Examples of each type are given in Supplementary Figure~\ref{fig:noise_example}. Each noise intervention was applied at five different intensities.

\subsection{Orientation and eccentricity tuning}
After training MNIST and CIFAR-10, we evaluated the responses of the trained models on a standard stimulus set typically used for retinotopic mapping of human occipital cortex. To study angular and orientation tuning we presented the trained networks with a rotating wedge (36 positions, angle extent $10^\circ$, $\text{radius} =14$ pixels). To study eccentricity tuning, we presented the network with ring images (13 different radius levels). For each unit in the grid this produced a 36-element series for the wedge angle and a 13-element series for ring eccentricity. 

\paragraph{Orientation analysis.} To describe angular tuning in topographic units, for each unit we measured responses to the 36 wedge stimuli. We applied a Fast Fourier Transform to each unit’s 36-point response profile and extracted the spectral power for the first five harmonics (cycles 1–5). These components reflect different angular periodicities: cycle 1 indicates preference for a single direction, cycle 2 for 180$^\circ$ symmetry consistent with orientation tuning, and cycle 4 indicates fourfold (90$\degree$) angular periodicity. We defined the \textit{dominant harmonic} for each unit as the harmonic (cycle 1-5) with the largest spectral power (excluding the DC component; i.e., the mean value of the signal). These dominant harmonic labels were assigned to the $11 \times 11$ grid.

% REMOVED from this version; redundant with Moran's
% To quantify the local spatial organization of harmonic preferences, we computed a \textit{neighborhood agreement score} for each unit. For a given unit, we identified its immediate spatial neighbors (up to 8 surrounding units) and calculated the proportion that shared the same dominant harmonic. The \textit{mean neighborhood agreement} was defined as the average of these proportions across all units in the grid, and the standard deviation reflected variability in local consistency.

\paragraph{Eccentricity analysis.}
To study eccentricity response profiles, we analyzed each unit's response to the 13 ring stimuli of increasing eccentricity by fitting a linear model to the 13 response values. Units with a Pearson correlation coefficient of $|r| > 0.8$, were labeled as \texttt{increasing} or \texttt{decreasing} depending on the sign of $r$.  For units not showing a linear profile, we evaluated if the response was selective to a particular eccentricity, indicating a band-pass response. To test this, we fitted a four-parameter Gaussian function (baseline, amplitude, center, width) to the unit's 13-point response profile. The quality of fit was determined using the coefficient of determination ($R^2$), and a unit was categorized as showing a \texttt{bandpass} response when $R^2 > 0.5$. Profiles that did not meet any of the above criteria were labeled \texttt{flat}.

\subsection{Functional localization}
For each trained model, we measured localization to understand how closely located were correlated units. We first computed the correlation between every pair of units’ activations across the test images. Given a chosen correlation threshold, we marked a pair of units as connected if their correlation exceeded that threshold, producing a binary connectivity matrix (connected vs. not connected). Next, for each unit, we identified all other connected units and computed the mean Euclidean distance to those. In a last step, we averaged these per-unit distance values across all units in the grid to obtain a single localization score for the model at that threshold. Smaller values indicate that strongly correlated units are more spatially clustered.

\subsection{Weight correlations and activation correlations}\label{sec:MethodWeightCorrel}
After WS and AS training, we computed, for each unit in the \(11\times 11\) grid, its average incoming-weight correlation with neighboring units. Incoming weights refer to the weight matrix from the preceding global-average pooling layer to the 121-unit grid layer. The correlation \(r_{ij}^{(\mathrm{In})}\) between the incoming weight vectors of units \(i\) and \(j\) was the Pearson correlation of the two vectors:
\begin{equation}
r_{ij}^{(\mathrm{In})}
=
\frac{\sum_k (w_{ik}-\bar{w}_i)(w_{jk}-\bar{w}_j)}
{\sqrt{\sum_k (w_{ik}-\bar{w}_i)^2 \; \sum_k (w_{jk}-\bar{w}_j)^2}} \, ,
\label{eq:weightR}
\end{equation}
where \(w_{ik}\) and \(w_{jk}\) denote the \(k\)-th afferent weight into units \(i\) and \(j\), respectively, and \(\bar{w}_i\) and \(\bar{w}_j\) are the means of their incoming weight vectors (dimension 64 for MNIST; 256 for CIFAR-10). For each unit \(i\), we then computed the mean correlation over its immediate (Moore) neighborhood \(S_i\). The neighborhood size \(|S_i|\) was 3, 5, or 8 for corner, edge, and interior units, respectively:
\begin{equation}
R_i^{(\mathrm{In})}
=
\frac{1}{|S_i|}
\sum_{j \in S_i} r_{ij}^{(\mathrm{In})} \, 
\label{eq:meanWeighCor}
\end{equation}

We evaluated activation correlations using the same logic described above for computing of incoming weight correlations. The difference was that for each unit-pair, we computed the correlation of their activation profiles for a minibatch of images. This allowed computing, for each unit, its average neighbourhood correlation. It also allowed studying the entire distribution of pairwise correlations. 

Both the AS and WS spatial loss terms operate locally, by encouraging each unit to be similar to its immediate neighbors in the grid. This differs from global objectives \citet{poli2023introducing}, which use a spatial loss that encourages activation similarity to decrease with grid distance across \textit{all} unit pairs. Global objectives penalize cases where highly correlated units are positioned far apart. They therefore discourage the formation of multiple, disconnected clusters of similarly-responding units, and instead produces spatially contiguous regions of response-similar units. For completeness, we also applied a global activation loss (as in \citet{poli2023introducing}), and found that it produces qualitatively different topographic activations than the local activation similarity loss (see Appendix Section \ref{sec:appendGlobalSim}).

\subsection{Expert unit analysis}
Following prior work \cite[e.g.,][]{soo2025expertUnits}, we define expert (class-selective) units whose activations reflect one-vs-rest discrimination of a target class. For each unit in the $11\times11$ topographic grid layer and each class, we quantified discriminability using the area under the receiver operating characteristic curve (AUC), computed from the unit’s post-ReLU activations over the test set. The AUC estimates the probability that the unit’s activation for an input from the target class exceeds activations for inputs from any other class.

We use two AUC thresholds. Units with $AUC > 0.70$ were classified as moderately class-selective; they produce higher activation to the target class in more than 70\% of random class–nonclass comparisons. Units with $AUC > 0.90$ were classified as strongly class-selective. For each condition, we counted the number of units meeting these criteria, and normalized by the total number of units ($n=121$) to obtain a proportion. Condition-level values were obtained by averaging the corresponding measures across the ten independently trained models for each condition.

To quantify the distribution of expert units over classes, each expert unit was assigned to the class for which it achieved its maximal AUC. This produced a distribution of expert units for each model family. We then computed the entropy of this distribution and normalized it by the maximum entropy for ten classes ($log10$). We refer to this quantity as \textit{expertise balance}; it equals 1.0 when expert units are uniformly distributed across classes and lower otherwise. 

Finally, we quantified the selectivity of expert units by computing, for each unit, the difference between its highest AUC (corresponding to the most discriminative class) and its second-highest AUC across classes. This AUC gap shows how exclusively a unit discriminates its preferred class relative to alternative classes. We refer to this as \textit{expertise selectivity}.

\section{Results}
\subsection{Accuracy}

\begin{figure}[]
    \centering
    \includegraphics[width=1\linewidth]{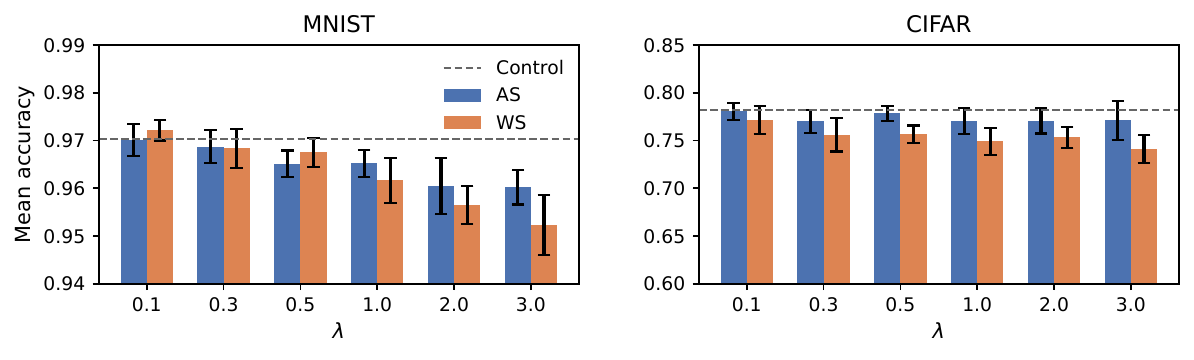}
    \caption{\textbf{Test accuracy as a function of topographic regularization strength $\bm{\lambda}$}. Mean classification accuracy for control (dashed), AS (blue), and WS (orange) models is shown for the different  values of $\lambda$.  Error bars indicate $\pm$ s.e.m.}
    \label{fig:accuracy}
\end{figure}

Control models were generally associated with the highest accuracy, followed by AS. Under weaker topographic regularization strength, AS and WS achieved similar accuracy to that of control. For MNIST, WS slightly outperformed the control models at the lowest lambda level (Figure~\ref{fig:accuracy}). Moderate topographic regularization produced a drop in accuracy up to 3\%, and more strongly for WS. This drop in accuracy was also observed in previous end-to-end topographic models \citep{margalit2024unifying, lu2025end, rathi2024topolm}.

\begin{comment}
We also evaluate how well the control, AS, and WS models were calibrated \citep{guo2017calibration}. The topographic models showed slightly less effective calibration as compared to the control model, but still very strongly differentiated between the top-1 and second best target (see Figure \ref{fig:mergedCalibration} in Appendix for details).    
\end{comment}

\subsection{Robustness}

\begin{figure}[]
  \centering
  \begin{tabular}{cc}
    \includegraphics[width=0.49\textwidth]{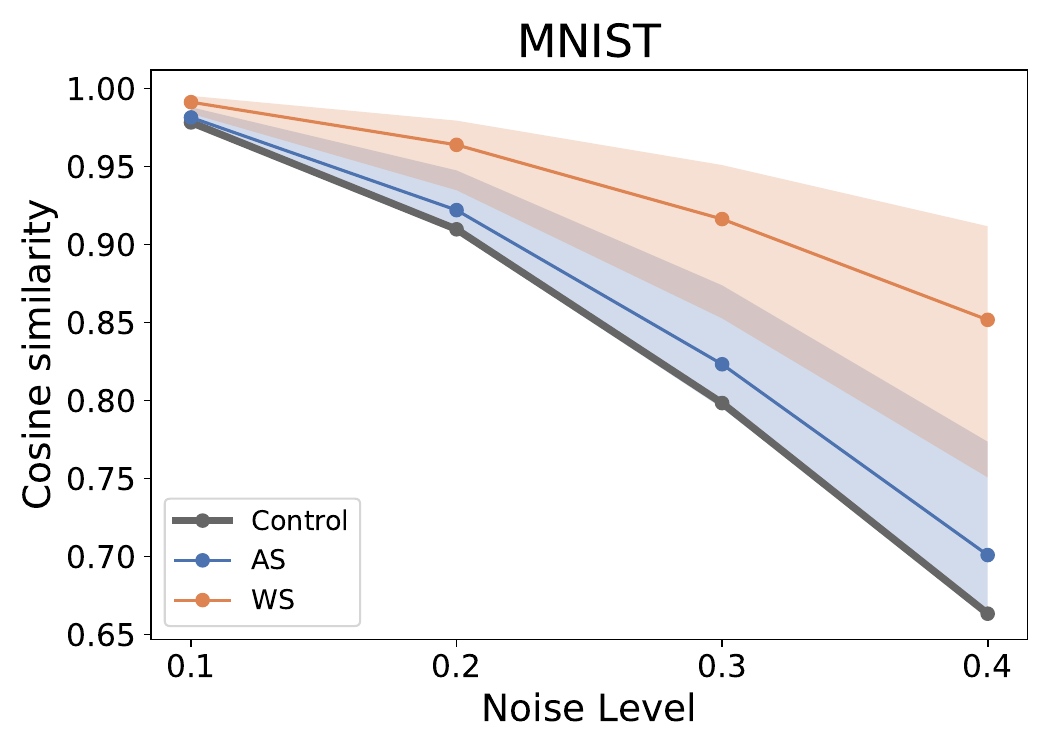} & \includegraphics[width=0.49\textwidth]{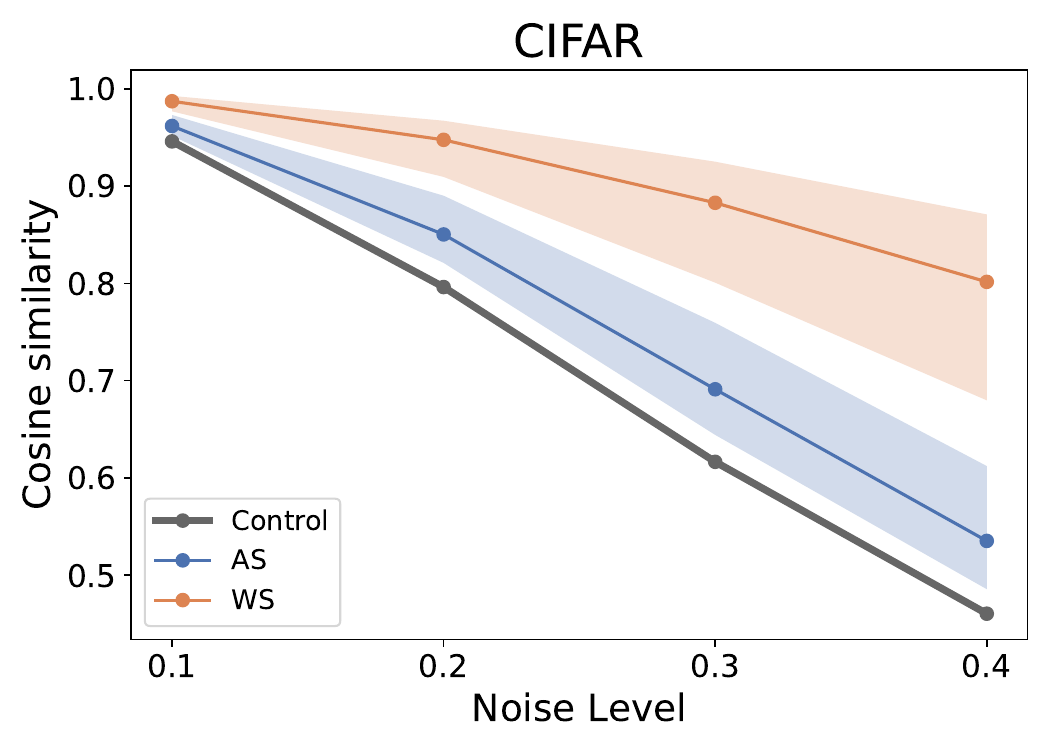} \\ 
    \includegraphics[width=0.49\textwidth]{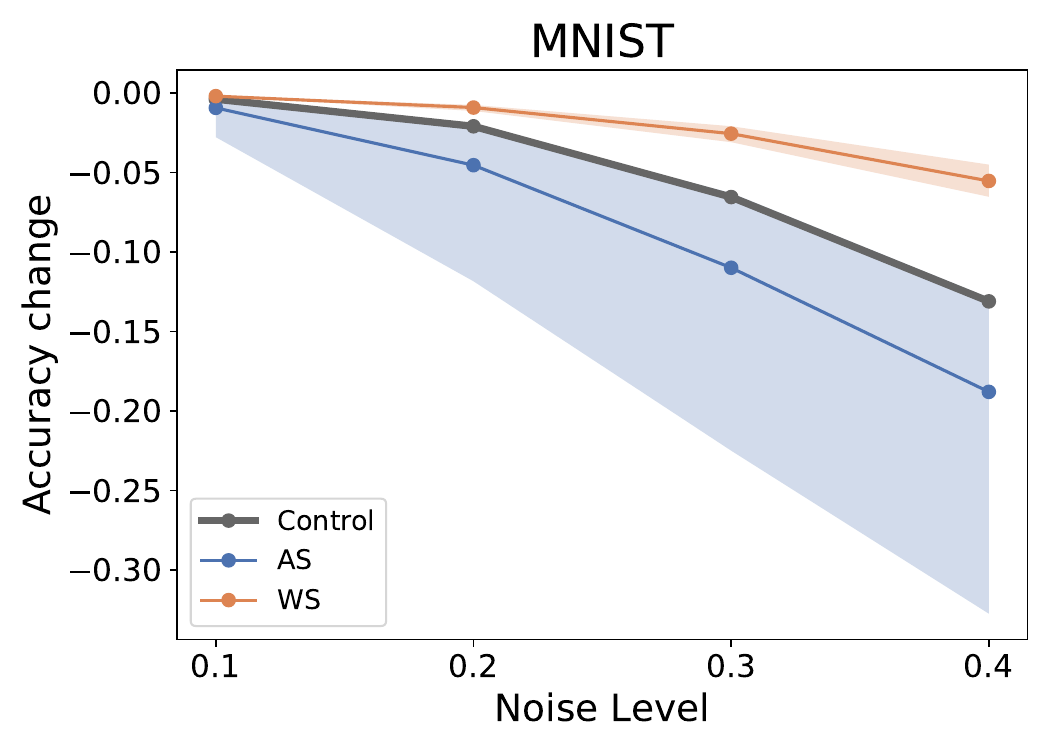} & \includegraphics[width=0.49\textwidth]{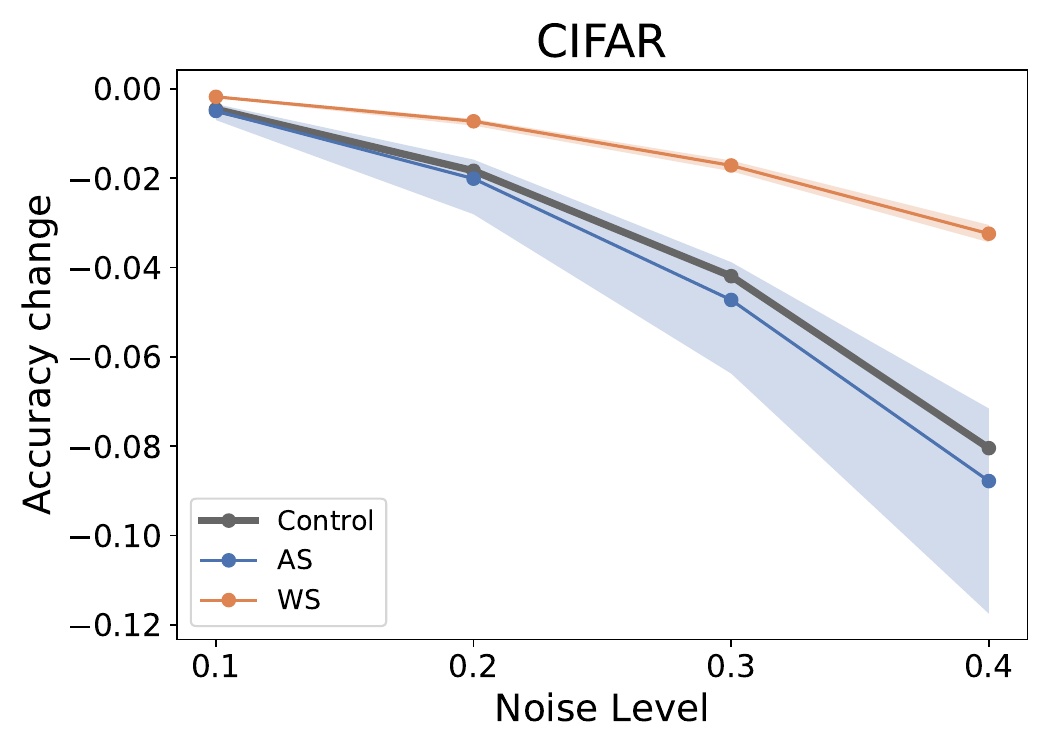} \\ 
  \end{tabular}
    \caption{\textbf{Robustness to weight perturbations as a function of noise levels.} Robustness is evaluated by changes in representational geometry (top row) and test accuracy (bottom row) for increasing levels of additive weight noise. Representational geometry is defined as the representational similarity matrix (RSM) computed from class weight vectors (CWVs) in the readout layer. Cosine similarity between original and perturbed representations is shown in the top row; corresponding changes in test accuracy relative to baseline (non-perturbed) models are shown in the bottom row. Shaded regions indicate min--max range across $\lambda$ levels. Values computed separately for each $\lambda$ level and model are reported in Supplementary Figure~\ref{fig:Approbweight}.}
    \label{fig:robweight}
\end{figure}

\paragraph{Robustness of representational geometry.}
We added Gaussian noise of varying magnitudes to the weight matrix connecting the topographic and classification layers and evaluated the resulting representation by constructing a representational similarity matrix (RSM) from the class-weight vectors. The intervention showed that WS training resulted in a more robust representational geometry than for AS and control models (see Figure~\ref{fig:robweight}). First, the second-order similarity between the baseline and perturbed RSMs was consistently higher for WS models, for both MNIST and CIFAR-10. WS models, particularly when trained with larger $\lambda$, were consistently top-ranked, showing less degradation in representational geometry. The control models showed the least robustness. %For example, on CIFAR-10, WS models achieved similarity values in the range of 0.7--0.9, whereas AS models ranged from 0.5--0.6. For MNIST, these values were 0.75--0.9 and 0.65--0.77, respectively.

Robustness of representation was also evident in the accuracy drop statistics: for MNIST, WS models showed a relatively small drop of 0--5\%, but AS models showed larger drops of 10--30\%. A similar pattern was observed for CIFAR-10, where the accuracy drop for AS was, on average, more than twice that of WS. In both datasets, WS outperformed the control models as well. These results suggest that WS training stabilizes representations under perturbation, which in turn also better maintain classification performance.

\paragraph{Robustness to image corruption}

\begin{figure}[]
  \centering
  \includegraphics[width=1\linewidth]{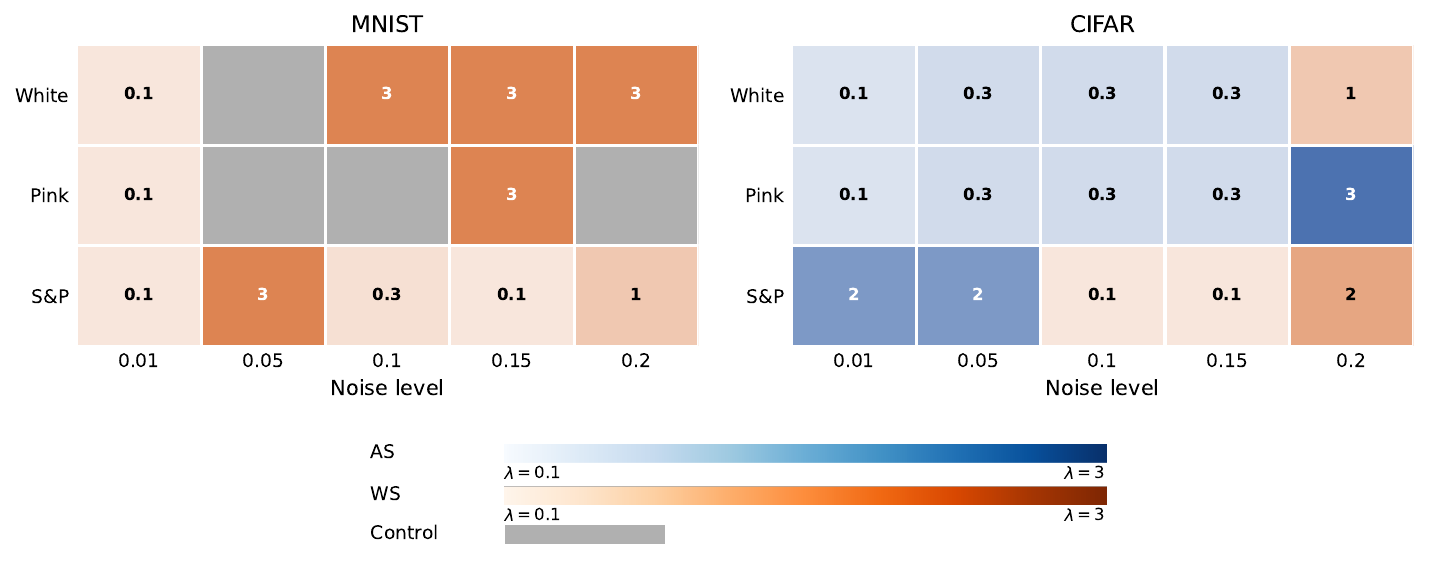}
    \caption{\textbf{Model-group robustness under input corruptions.} For each dataset and corruption type (white, pink, and salt-and-pepper), the most robust model group is defined as the one showing the smallest drop in test accuracy relative to the uncorrupted baseline. On MNIST, WS models are most often the most robust; on CIFAR-10, AS models are generally most robust. Each cell reports the $\lambda$ value associated with the winning group. A full breakdown of accuracy drops across $\lambda$ and corruption intensities is reported in Supplementary Figure~\ref{fig:Appnoise_acc}.}
  \label{fig:noiseMainFigu}
\end{figure}

Figure~\ref{fig:noiseMainFigu} shows, for each level and type of noise, which model family (AS, WS, control) was most robust to input corruption.  For MNIST, WS tended to be most robust, and for CIFAR-10, AS tended to be the most robust. (Figure~\ref{fig:Appnoise_acc} presents the full breakdown, presenting baseline-normalized performance by noise type, noise level and $\lambda$.)

It should be noted that at the lowest regularization strength ($\lambda = 0.1$), AS and WS showed stronger robustness to input corruption and stronger representational robustness than control models, but similar classification accuracy. Prior studies have tried training for robustness, finding this typically reduces accuracy \citep{hendrycks2019benchmarking, lopes2019improving, tsipras2018robustness, su2018robustness}. Our findings suggest that moderate topographic regularization can achieve a similar aim without a strong trade-off.

% \textcolor{blue}{Regarding the relationship between accuracy and robustness, there is no guarantee that higher accuracy translates into greater robustness \citep{hendrycks2019benchmarking, lopes2019improving, tsipras2018robustness, su2018robustness}, and our results are consistent with this.}

% shows the impact of image corruption on classification accuracy, for different levels of noise and levels of $\lambda$. All graphs show baseline-normalized activity. Panels in the figure show the impact of white noise, pink noise and salt-and-pepper noise, respectively. In both datasets, accuracy degraded as the noise level increased. In MNIST, the WS models were generally more robust than the AS models across most of the noise levels and the strength of the spatial constraints. In CIFAR-10, the WS models only demonstrated increased robustness for higher noise levels, but were otherwise on par with AS. Among the three types of noises, salt-and-pepper noise showed the largest gap between the WS and AS. In case of very high spatial constraint ($\lambda = 3$), the WS models either approximated or even surpassed the performance of the control non-topographic models.

% \begin{figure}[]
%   \centering
%   \fbox{\parbox[c][4cm][c]{0.8\linewidth}{\centering Figure placeholder}}
%   \caption{\textit{WS models are generally more robust to input noise than AS models, especially for high noise levels.}}
%   \label{fig:placeholder}
% \end{figure}

\subsection{Activation entropy, sparsity and effective dimensionality}

\begin{figure}[]
    \includegraphics[width=1\linewidth]{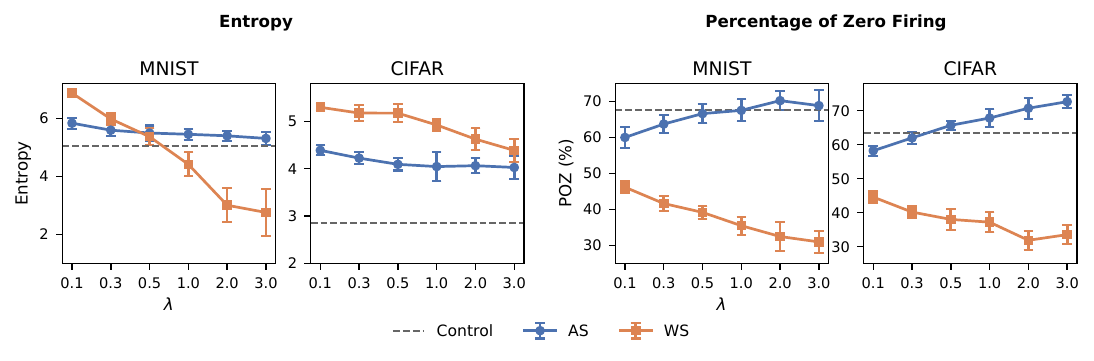}
    \caption{\textbf{Unit-level entropy and Percentage-of-Zero activations as a function of topographic regularization strength ($\lambda$)}. The two left plots show average entropy of pre-ReLU unit activations for MNIST and CIFAR-10 across values of $\lambda$. The two right plots show the average Percentage-of-Zero (PoZ) of post-ReLU unit activations. Error bars indicate $\pm$ s.e.m.}
    \label{fig:EntropyPoz}
\end{figure}

The variance or entropy of a unit’s activations is often treated as an indicator of its functional importance within an ANN. Higher entropy reflects greater sensitivity to input variation, and lower-entropy elements are targets for pruning \citep{polyak2015channel, wang2021filter, luo2017entropy, liao2023can, lu2025end}. Similarly, a unit's tendency to remain inactive across inputs, quantified as PoZ (Percentage of Zeros), is used as a pruning criterion, with higher PoZ indicating less functional importance \citep{hu2016network}. We analyzed the entropy and PoZ of the grid units in AS, WS, and control models. Entropy was computed from each unit’s pre-ReLU activations across the  10,000 images test images. PoZ was computed post-ReLU, reflecting the fraction of inputs for which a unit’s activation was zero. We computed the average entropy and PoZ across units within each grid. As shown in Figure~\ref{fig:EntropyPoz}, WS models tended to present higher entropy and lower PoZ than both AS and control models. For MNIST, the advantage held at lower levels of $\lambda$, and for CIFAR-10, the advantage held across all  $\lambda$ levels. This suggests that WS training produces a larger number of responsive units, which also better differentiate the input images. 

\begin{figure}[]
  \centering
    \includegraphics[width=\textwidth]{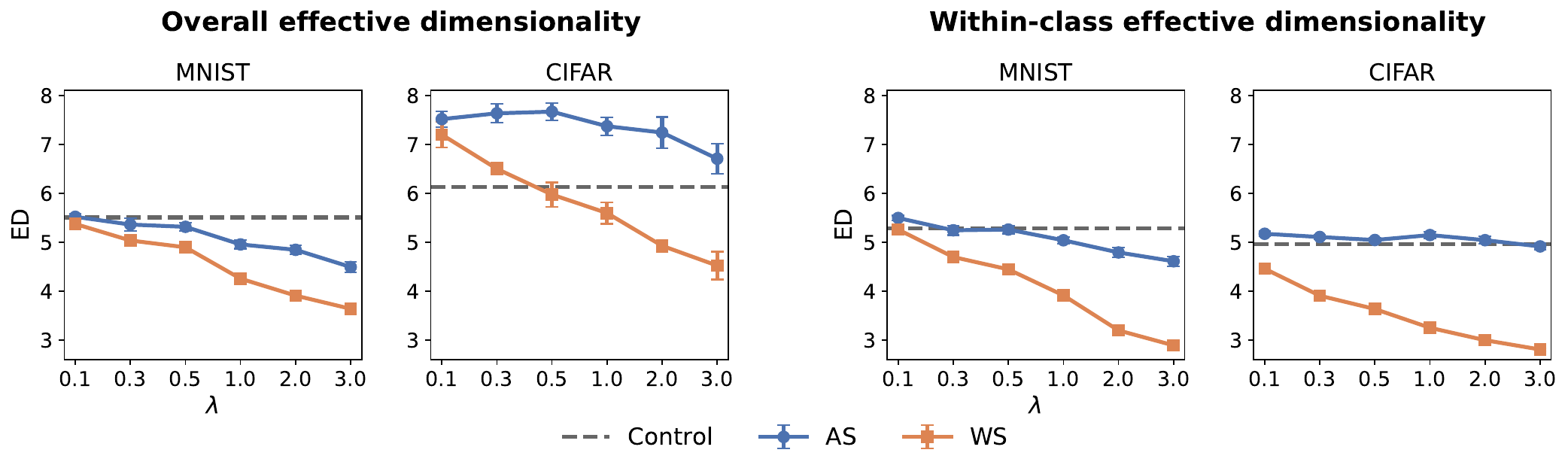}
  \caption{\textbf{Effective dimensionality of topographic representations}. Overall effective dimensionality (ED; left) and within-class effective dimensionality (right) of the fc1 layer as a function of the spatial regularization strength $\lambda$. Overall ED is computed from the activation covariance across all images; within-class ED is computed for each class separately and then averaged across classes. Error bars denote s.e.m, and dashed lines indicate control models.}
  \label{fig:effectiveDim}
\end{figure}

To determine how these unit-level properties translate into the organization of information in the latent space, we examined the Effective Dimensionality (ED) of model activations \citep[e.g,][]{margalit2024unifying, qian2024local, deb2025toponets}. Using the activations evoked by the 10,000 test images, we computed ED from the covariance matrix of all images to estimate the overall dimensionality of the representation. In addition, we computed within-class ED by isolating activations for each class and averaging ED across classes. We find that AS and WS models produced different effects on the dimensionality of latent representations (Figure \ref{fig:effectiveDim}). For both datasets, AS models were associated with higher overall effective dimensionality than WS and control models, meaning that information resided in a larger number of latent dimensions. These results suggest that WS training produces a more redundant, low-dimensional class representations. As expected, increasing $\lambda$ led to a gradual reduction in both overall and within-class effective dimensionality, though the trend was stronger for WS. We also note that for CIFAR-10, both AS and WS produced higher overall effective dimensionality than control models at the lower $\lambda$ levels. This shows that stronger local correlations do not necessarily result in reduced dimensionality. 

\subsection{Functional localization metrics}
\subsubsection{Functional co-localization}
\begin{figure}[!t]
    \centering
    \includegraphics[width=1\linewidth]{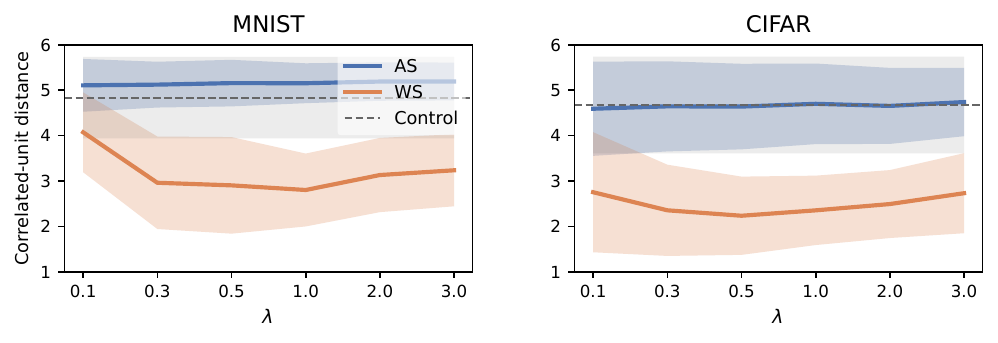}
    \caption{\textbf{Correlated-unit distance in the topographic grid as a function of topographic regularization strength ($\lambda$)}. Mean spatial distance between pairs of units whose activity-correlation exceeds a threshold $\alpha \in {0.1, 0.3, 0.5, 0.6, 0.7, 0.8}$. Solid lines indicate distances averaged across correlation thresholds $\alpha$; shaded regions denote $\pm$1 SD across $\alpha$ thresholds. Distances computed separately for each correlation threshold and model are reported in Supplementary Figure~\ref{fig:Appconnect_dist}.}
    \label{fig:connect_dist}
\end{figure}

To evaluate to what extent units with similar firing patterns were positioned closely on the topographic grid, we defined two units as belonging to the same functional network if their activation patterns exceeded a correlation threshold $\alpha$. We then computed the average Euclidean distance between connected units (see Methods). Figure \ref{fig:connect_dist} shows the results for WS and AS (for simplicity, the figure presents the mean and $\pm 1SD$ over responses of $\alpha$; details are presented in Supplementary Figure~\ref{fig:Appconnect_dist}). As the figure shows, for both MNIST and CIFAR-10, WS was associated with smaller distances between correlated units, and a detailed analysis shows this held for all levels of $\alpha$.

% Another finding evident in the figure is that increasing the strength of the spatial constraint $\lambda$ did not produce a monotonic decrease in distances. For WS, the shortest distances were found for $\lambda$ levels of 0.3 or 0.5, with distances increasing rather than decreasing at higher levels.  For AS, the level of $\lambda$ had a weaker effect, with the lines remaining relatively flat. 

% We also analyzed data for the control condition, by assigning unit indices to the grid randomly.  Average distances in the control condition were qualitatively quite similar to those in AS. For $\alpha = 0.1$ the distances for AS were very slightly below the control (highest AS value 5.64; control, 5.56), and the same held for $\alpha = 0.3$ (highest AS value 5.57; control, 5.58). However, for the other levels of $\alpha$, the distances for the control condition were always smaller than AS. In contrast, the distances for the control condition always strongly exceeded those of WS.

% The distances for AS were larger than those of WS. For WS, at $\alpha = 0.1, 0.3, 0.5, 0.6$, we again find a U-shaped result pattern. However, for the two highest thresholds, there was a monotonic increase in distance as a function of $\lambda$. 

% For AS, the distances were always lower than the control for $\alpha = 0.1, 0.3, 0.5, 0.6$ , independent of the level of $\lambda$. For $\alpha = 0.7$, AS distances were below control in all cases, apart from $\lambda = 0.3$, while for $\alpha = 0.8$ AS distances were always higher than control distances. 

%To summarize, WS training produced the shortest distances between similarly-activating units, suggesting stronger functional organization than that found in AS. Interestingly, for WS, increased smoothing constraints produced a U-shaped pattern for higher levels of $\alpha$. This suggests that as the spatial constraint increases, it initially limits strongly similar activations to the each unit's immediate neighborhood, but then develops stronger regional homogeneity as the constraint increases. 

\subsubsection{Spatial autocorrelation of unit activations}

\begin{figure}[]
    \centering
    % ---------- First image (2/3 page width) ----------
    \begin{minipage}[t]{0.60\textwidth}
        \centering
        \includegraphics[width=\linewidth]{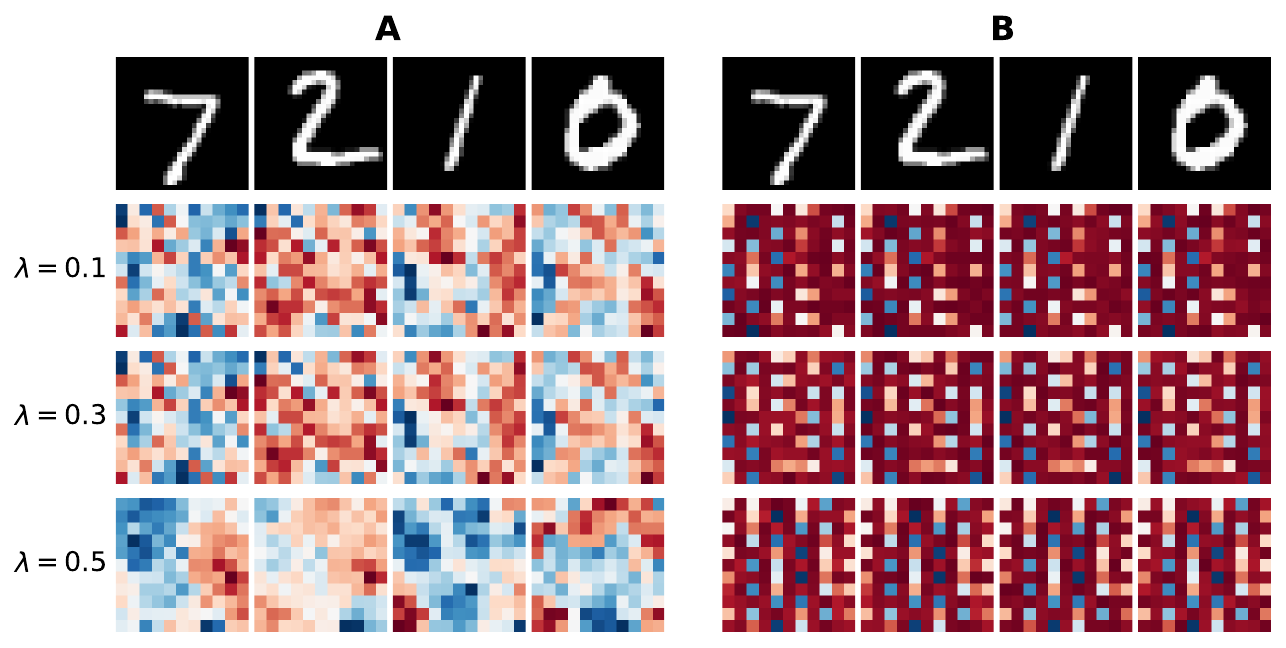}
        % \captionof{subfigure}{Main comparison}
    \end{minipage}
    \hfill
    % ---------- Second image (1/3 page width) ----------
    \begin{minipage}[t]{0.35\textwidth}
        \centering
        \includegraphics[width=\linewidth]{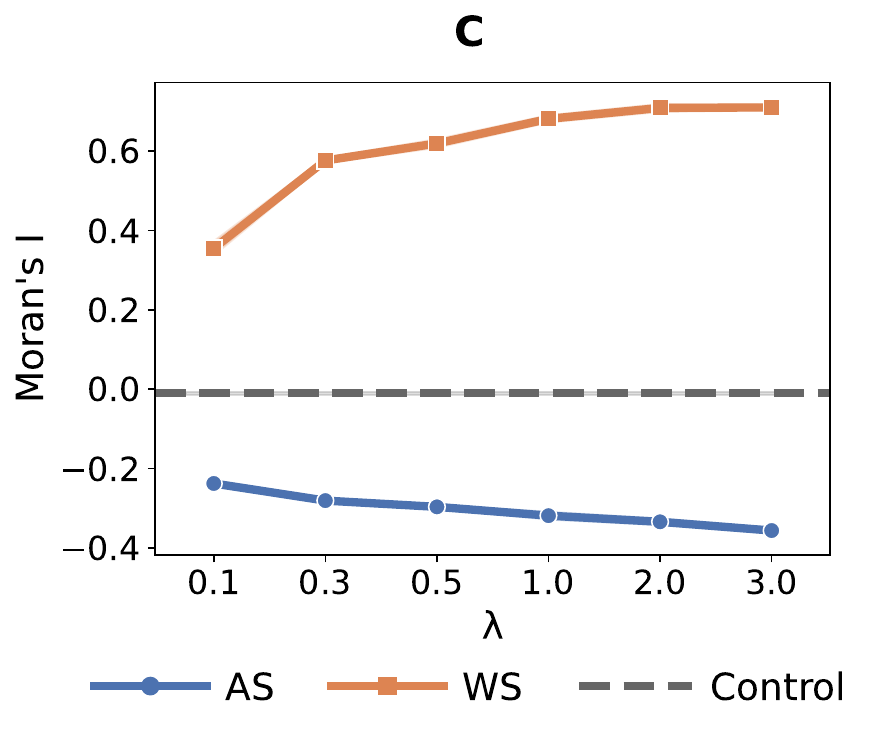}
        % \captionof{subfigure}{Auxiliary result}
    \end{minipage}

    \caption{\textbf{Spatial smoothness of activation maps under activation- and weight-similarity training}.  Sample activation maps from a topographic grid layer are shown for WS-trained models (Panel A) and AS-trained models (Panel B) at three values of $\lambda$. WS training is associated with spatially smooth activation patterns, with smoothness increasing as $\lambda$ increases.  AS training produced alternating patterns of correlated and uncorrelated units. Panel C shows Moran’s I, which quantifies spatial autocorrelation of grid activations. Similar patterns are observed for CIFAR (Appendix Figure \ref{fig:Appactivation_comparison_cifar}).}
    \label{fig:activation_comparison}
\end{figure}

To understand the spatial organization produced by WS and AS training, we quantify activation smoothness, weight similarity, and activation correlations among neighboring and non-neighboring units on the topographic grid. We first quantified the spatial smoothness of activation maps using Moran's \textit{I}, which is a spatial smoothness statistic \cite{moran1950notes}, previously used in related work \cite[e.g.,][]{rathi2024topolm}. Positive values indicate smoother transitions among neighboring units, negative values indicate spatial dispersion (e.g., high-low activation transitions between  neighbors), and zero indicates randomness. We applied this metric to the pre-ReLU activation maps produced by each image in each model, and averaged within AS, WS, and control models. Figure~\ref{fig:activation_comparison} presents the results for MNIST, and almost identical findings were found for CIFAR-10 (see Supplementary Figure~\ref{fig:Appactivation_comparison_cifar}).

As Figure~\ref{fig:activation_comparison}C shows, WS models produced consistently positive Moran's \textit{I} values, which increased monotonically with $\lambda$. In contrast, the AS models produced negative scores, reflecting transitions between high and low activation values. The control models approximated zero as expected. 

To visually appreciate the distribution of local activation similarity, Figure \ref{fig:activation_comparison}A shows activation maps from three randomly selected WS models trained under regularization strength of $\lambda = 0.1, 0.3, 0.5$. It is evident that WS training produces substantial spatial autocorrelation even at $\lambda = 0.1$.  Unlike WS, AS training (Figure~\ref{fig:activation_comparison}B) did not produce spatially smooth activation maps. Instead, it produced a large number of adjacent units with very similar activations, mixed with units showing very dissimilar activations, resulting in a striped-like activation pattern. This pattern is clarified by the correlation analysis reported below.

\begin{figure}[]
    \center
    \includegraphics[width=1\textwidth]{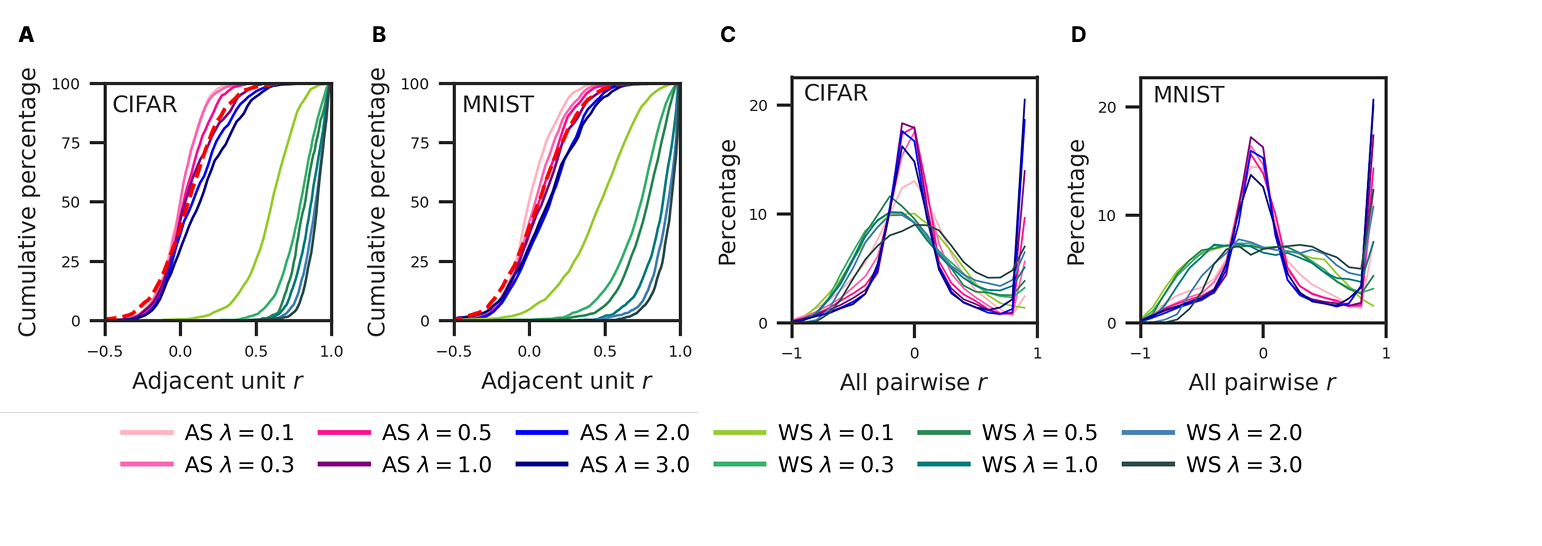}
    \caption{\textbf{Different patterns of correlated activations under activation- and weight-similarity training}.  Cumulative distributions of pairwise activation correlations between adjacent units are shown for CIFAR-10 (Panel A) and MNIST (Panel B). Distributions of correlations computed across all unit pairs are shown for CIFAR-10 and MNIST (Panels C and D). Neighborhood correlations (Panels A and B) are stronger under WS training. Correlations computed across all unit pairs (Panels C and D) show a larger proportion of near-perfect correlations under AS training.}
    \label{fig:comppropActCorrel}
\end{figure}

To understand these patterns, we analyzed activation correlations at both local and global scales. At the neighborhood level, WS training produced substantially stronger correlations between adjacent units than AS across all values of $\lambda$, even though its objective did not encourage activation similarity (Figure \ref{fig:comppropActCorrel}, A–B). In AS models, higher $\lambda$ increased neighborhood correlations relative to control models, but lower $\lambda$ shifted the distribution leftward, indicating reduced local coherence. At the level of all unit pairs (Figure \ref{fig:comppropActCorrel}, C–D), AS and WS training produced qualitatively different correlation structures: AS produced a bimodal distribution with one peak near zero and another at near-perfect correlation ($r\approx1$). This means that the AS objective was achieved by strongly coupling a subset of units while leaving others weakly correlated. WS, in contrast, produced a flatter distribution with higher correlations across a broader range of unit pairs. Both AS and WS increased local similarity among incoming weight vectors relative to control (Supplementary Figure \ref{fig:bothmodels_incomingWeights}), though more strongly for WS as would be expected from its training objective. 

\subsection{Reorganization of angular and eccentricity tuning under topography}

\begin{figure}[]
    \centering
    \includegraphics[width=1\linewidth]{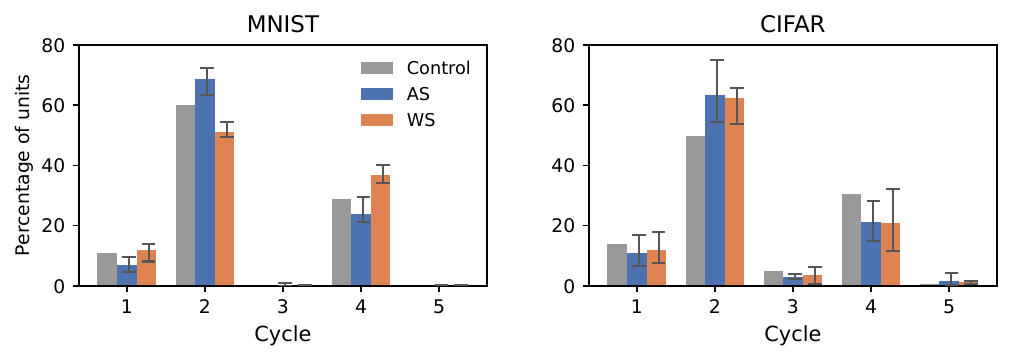}
    \caption{\textbf{Distribution of angular tuning profiles relative to control models under topographic training}. Percentage of units associated with each angular sensitivity profile (Cycles 1–5) is shown. Cycle 1: classic polar-angle tuning; Cycle 2: orientation-like tuning; Cycles 3–5: higher angular frequencies. Bars indicate mean percentages averaged across the six values of the regularization parameter $\lambda$; vertical whiskers indicate the full min–max range across $\lambda$. In both datasets, orientation-like tuning (Cycle 2) is the most prevalent profile, and AS and WS models differ from control models in the relative distribution of units across profiles. Full cycle-response data, shown separately for each value of $\lambda$, are reported in Figure~\ref{fig:AppharmonicsCIFARMNIST}.}
    \label{fig:harmonicsCIFARMNIST}
\end{figure}

Relative to control models, topographic regularization resulted in changes to tuning profiles. The strongest effects were found when analyzing orientation-like angular responses and for the balance between central and peripheral eccentricity coding. For angular tuning, topographic regularization changed the balance between orientation tuning (\texttt{cycle=2}) and symmetry tuning (\texttt{cycle=4}) responses (Figure~\ref{fig:harmonicsCIFARMNIST}). For CIFAR-10, both AS and WS increased the prevalence of orientation tuning and reduced the number of units displaying symmetry tuning. For MNIST, there were also differences between AS and WS: AS increased the prevalence of orientation tuning already evident in the control models, but WS shifted units from orientation to symmetry tuning. Thus, topographic training  reweighted angular harmonics, but this  depended on dataset properties (MNIST vs. CIFAR-10) and the form of topographic regularization. Supplementary Figure~\ref{fig:AppWedgeHarmonics} shows the spatial distribution of angular tuning profiles in the topographic grid for WS-trained models.

\begin{figure}[]
    \centering
    \includegraphics[width=1\linewidth]{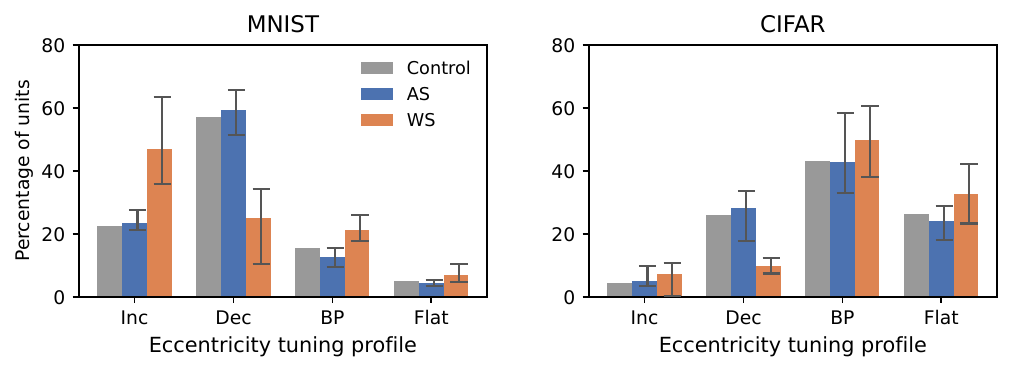}
    \caption{\textbf{Distribution of eccentricity tuning profiles relative to control models under topographic training}.  Percentage of units showing increasing (Inc), decreasing (Dec), band-pass (BP), or flat eccentricity tuning profiles is shown. Bars indicate mean percentages averaged across values of the regularization parameter $\lambda$; vertical whiskers indicate the full min–max range across $\lambda$. In both datasets, AS and WS models differ from control in the relative distribution of units across eccentricity tuning profiles. Eccentricity data, shown separately for each value of $\lambda$, are reported in Figure~\ref{fig:Appeccentricity_allprofiles}.}
\label{fig:eccentricity_allprofiles}
\end{figure}

A reorganization was also observed for eccentricity tuning (Figure~\ref{fig:eccentricity_allprofiles}). Here, tuning profiles reflect radial gain functions: \textit{decreasing} responses correspond to filters that weight central locations more strongly; \textit{increasing} responses weight peripheral locations more strongly. The most salient effect was that WS strongly reduced centrally preferring (decreasing) responses as compared to both control and AS. In MNIST, this reduction was accompanied by an increase in monotonically increasing responses, indicating greater sensitivity to peripheral locations. In CIFAR-10, where eccentricity is supposedly a weaker cue for categorization, WS instead produced an increase in band-pass responses. In summary, WS consistently de-emphasized central vision.

\subsection{Expert units}
 
\begin{figure}[]
    \centering
    \includegraphics[width=\linewidth]{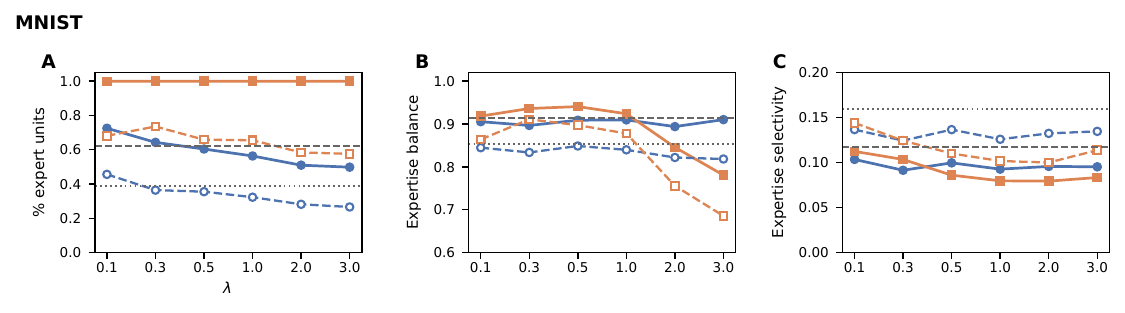}
    \includegraphics[width=\linewidth]{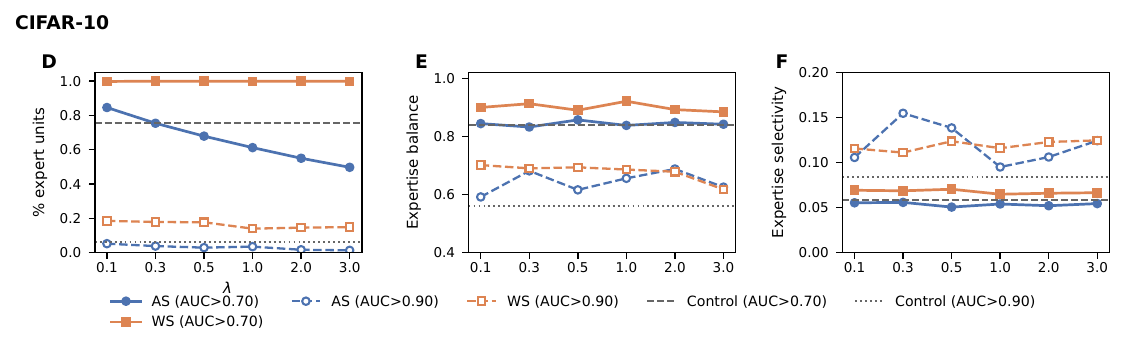}
  \caption{\textbf{Expert unit prevalence, balance, and selectivity}. Expert unit prevalence, defined as the proportion of units whose category discriminability exceeds a threshold (AUC $>0.70$ or AUC $>0.90$), is shown for MNIST (Panels A) and CIFAR-10 (Panels D). Expertise balance, quantified as the normalized entropy of expert units across categories, is shown in Panels B and E. Expertise selectivity, defined as the difference between the highest and second-highest AUC values for each unit, is shown in Panels C and F. See Methods for formal definitions of all measures.}
    \label{fig:experts}
\end{figure}

All models developed expert units that discriminated specific categories. We defined two levels of expertise: moderate-expertise units that systematically showed higher activity for one category in more than 70\% of random category–noncategory comparisons, and high-expertise units, corresponding to a 90\% threshold. The main finding, as shown in Figure \ref{fig:experts}, was that for both MNIST (Panel A) and CIFAR-10 (Panel D), WS models produced a larger proportion of both moderate- and high-expertise units as compared to AS and control. For AS, but not WS, the proportion of experts decreased with $\lambda$. 

We next examined how expert units were distributed across categories by quantifying the balance of expertise (Figure \ref{fig:experts}, B, E). Here, the results depended on the dataset.  For CIFAR-10, balance was quite similar for AS and WS, with both model families showing slightly higher balance than control models. For MNIST, the WS distribution of experts became less balanced at highest levels of $\lambda$, but for CIFAR-10, balance remained stable under $\lambda$.

We also studied the specific association of expert units with different categories (see Supplementary Figure \ref{fig:app_expert_distrib}). We found that topographic training shifted the distribution of experts. For MNIST, in control models most experts were dedicated to the digit \textit{3} followed by \textit{1,4,6,8}. For WS, a greater proportion of experts was allocated to \textit{4, 7}.  For CIFAR-10, in control models most experts were allocated to the \textit{automobile} category. For the topographic networks, this category was prominent (particularly for WS), but in addition, WS experts were allocated to \textit{truck, ship, airplane, horse and frog} but not to \textit{dog, deer, cat, bird}. 

Finally, we evaluated the selectivity of expert units by isolating expert units and quantifying the difference between the response to the most preferred category and the second-most preferred category (Figure \ref{fig:experts}, panels C and F).  We find that for both MNIST and CIFAR-10 the relative selectivity of expert units did not differ strongly among conditions. Taken together, the results show that topographic training strongly impacted the proportion and specificity of expert units. However, this depended on an interaction of dataset properties (MNIST vs. CIFAR-10) and the topographic loss itself.  

\section{Discussion}

We studied how topographic regularization impacts robustness and representational characteristics of shallow CNNs when applied during end-to-end training. We focused on two local topographic losses: activation similarity (AS), which encourages high Pearson correlation between neighboring units’ activation vectors (computed across inputs), and weight similarity (WS), which encourages similarity among adjacent units' incoming (afferent) weight vectors. Our main objectives were to evaluate (1) robustness to readout-weight perturbations and input corruptions, and (2) representational properties including the entropy, sparsity, similarity profiles of the weights and activations, effective dimensionality, the emergence of class-selective (expert) units, and the spatial organization of responses on the 2D grid including smoothness and functional localization. 

The following main findings emerge: (1) Overall, AS and WS models produced greater robustness to input corruption as compared to control models; (2) WS models were more robust to perturbations of the readout (classifier) weight matrix, accompanied by higher activation entropy and lower activation sparsity (lower PoZ); (3) WS models  produced better spatial organization, with stronger functional localization and smoother spatial variation on the grid; (4) Both AS and WS changed the distribution of pairwise correlations, consistent with the objectives of the two regularizer, but AS did not produce smooth activation maps on the 2D grid.

\subsection{Topographic regularization improves robustness to noise}
One key finding emerging from our study is that topographic regularization can produce models that are more robust to both weight perturbations and input corruptions. This is important because several approaches for achieving robustness against noisy inputs have  used training on corrupted data \citep[e.g.,][]{sietsma1991noiseTrain, lopes2019improving}. In prior work, robustness against parameter perturbations \citep{cheney2017robustness, arechiga2018effect, savva2023robustness} typically relies on specifically tailored loss terms \citep{tsai2021weightperturb}. In contrast, we find that robustness can be a byproduct of the topographic regularization itself. This suggests that a biologically inspired regularizer can be a viable strategy for improving robustness (see \cite{liu2023comprehensive} for review on robustness methods). Specifically, for both CIFAR-10 and MNIST, WS models showed advantages over AS and control models in terms of robustness to weight perturbation. In particular, representational similarity matrices of WS-trained models were more stable under weight perturbation, and classification accuracy degraded less. Similarly, WS models were often more robust against image corruption for MNIST, and AS models were often more robust for CIFAR-10. These results extend prior work showing that certain types of topographic architectures are more robust against adversarial attacks \citep{qian2024local, bashivan2025learning}.

Importantly, we observe a non-monotonic relationship between robustness and model performance.  For MNIST, WS models trained with a weak regularization ($\lambda = 0.1$) produced higher accuracy than control models (Figure~\ref{fig:accuracy}, MNIST). Notably, the same setting also produced stronger robustness than controls. This pattern contrasts with prior work reporting that improved accuracy is associated with reduced robustness, particularly in adversarial-robustness settings \citep{su2018robustness, tsipras2018robustness}. Our findings point to a potentially important exception, where a moderate topographic regularization can, under some settings, improve both classification performance and robustness.

Our findings also show that the relative accuracy of topographic models and non-topographic controls depends on regularization strength ($\lambda$). In fact, previous studies report mixed findings, with some reporting reduced accuracy \citep{margalit2024unifying, lu2025end, rathi2024topolm} or slightly improved performance \citep{poli2023introducing, zhang2025brain, deb2025toponets, rathi2024topolm} depending on setting. These spatial loss functions differed in their topographic objective, including local similarity between unit-weights or activations \citep{lu2025end, deb2025toponets},and global distance-weighted similarity penalties on activations \citep{margalit2024unifying, poli2023introducing} or weight magnitudes \citep{zhang2025brain, jacobs1992computational}. While the literature on this point is mixed, our findings suggest that the strength of topographic regularization is a key determinant.

\subsection{Topographic regularization reshapes model representations}
As noted in the introduction, in machine-learning practice, the presence of correlations among unit activations or among incoming weight vectors is often discouraged due to reduced feature diversity \citep{zbontar2021barlow, wang2020mma} and because models often aim to de-correlate representations \citep{cogswell2015reducing, rodriguez2016regularizing, jin2020does}. This emphasis differs from the role of correlations in biological systems, where correlated activity can serve to amplify specific signals \citep[e.g.,][]{shadlen1998variable}. In artificial networks, however, similar effects can be achieved without correlated populations, for instance, by using high downstream readout weights. Still, some machine-learning approaches intentionally produce structured redundancy, for example, by developing methods to cluster similar weights or activations into groups that enforce within-group similarity, with the intention of providing a basis for subsequent model compression \citep{han2015deep, zhang2018learning, neill2020compressing, wen2016discriminative}. The results of the AS condition are most relevant to this literature as we find that it produces a substantial proportion of units with very similar activation profiles, suggesting that it might be a useful loss term for subsequent pruning or compression. While we have not specifically studied the pruning or compression potential of topographic models, several studies have already shown that topographic models are more amenable to pruning \citep{poli2023introducing, deb2025toponets, lu2025end}.

The changes we find in representational organization were accompanied by changes to single-unit responses properties. Specifically, WS training was associated with higher unit entropy at the lower $\lambda$ levels and with lower PoZ. Higher entropy indicates greater response variability across inputs. The lower PoZ suggests reduced sparsity because units are active for a larger fraction of inputs. Both statistics depended on the strength of regularization, and differed for AS and WS models. In particular, increasing $\lambda$ produced a gradual decrease in PoZ for WS models but a gradual increase in PoZ for AS models. 

Consistent with prior reports \citep{margalit2024unifying, qian2024local, deb2025toponets}, we found that the topographic regularizer could reduce the overall effective dimensionality of the representations as $\lambda$ increases (Figure \ref{fig:effectiveDim}). WS training tended to produce lower effective dimensionality than both AS and control models. However, not all topographic conditions were associated with lower overall effective dimensionality; specifically, for CIFAR-10, both AS and WS showed higher dimensionality than control for the two lower levels of regularization, and AS models exceeded control for all $\lambda$ levels. In contrast, \textit{within-class} effective dimensionality was consistently at or below control levels. This suggests that the topographic regularization mainly reduces within-class variation.

Lower-dimensional representations have been linked to increased robustness, as they are more resilient to various types of perturbations \citep{sanyal2018robustness, awasthi2020adversarial}. Related findings in non-topographic models also suggest that encouraging similarity in learned features or activations improves robustness \citep{nassar2021noise, gourtani2024improving}. Importantly, our results (Figure~\ref{fig:effectiveDim}) suggest that robustness is not necessarily accompanied by lower dimensionality: in some cases, topographic regularization was associated with both higher overall effective dimensionality and greater robustness.

When analyzing the spatial organization of activity, we found that WS models produced higher spatial smoothness (Moran's \textit{I}) and shorter spatial distances between highly correlated units. AS models, in contrast, produced a larger proportion of high-correlation unit pairs, but did not increase spatial smoothness. Instead, AS introduced a striped-like organization of activation, with lower smoothness than found for control models. We interpret this to be a consequence of the local scope of our AS objective: the model could minimize the local loss by producing neighbor pairs with very high correlations, without enforcing smooth variation across the full grid. This finding was not reported in prior work using activation-similarity objectives, which implemented \textit{global} losses that penalize high similarity at large spatial distances \citep{poli2023introducing, margalit2024unifying, rathi2024topolm}. Consistent with this, when we implemented a global activation-similarity loss, we found that we could recover prior findings showing that the loss term produces spatial smoothing on the grid (Appendix \ref{sec:appendGlobalSim}).

Interestingly, AS and WS also produced different angular and eccentricity tuning profiles. The clearest example was the eccentricity tuning patterns produced by MNIST training. Here, WS models produced an eccentricity preference profile in which more peripheral ring locations were associated with stronger responses. In contrast, AS training produced the opposite bias, with stronger responses for more central locations. Another example was that for MNIST, WS training produced fewer units showing orientation tuning (\texttt{cycle=2}) than AS, but more units showing symmetry tuning (\texttt{cycle=4}). Both regularizers increase local correlations, but only WS produced smooth transitions and functional clustering. 

Regarding the emergence of expert units, we find that topographic regularization influences unit-level specialization by changing the number and distribution of class-selective expert units. The difference in the proportion of expert units between WS and AS, together with the decreasing prevalence of experts as $\lambda$ increased in AS models, suggests that both the type of loss function and the regularization strength control the production of category-selective elements. The AUC gap between the top and second-best class was similar across conditions, suggesting that topographic regularization mainly affects expert prevalence and class coverage rather than increasing unit selectivity.

\subsection{Implications and future directions}
An important implication for machine-learning applications is that moderate topographic regularization strengths can improve both task performance and robustness to noise. This suggests that modulating local correlations may be beneficial in some settings, even when brain-like spatial smoothness is not an objective in itself. This can be simply implemented by adding a topographic regularizer to the training objective. From a biological perspective, our findings suggest that natural topographic constraints that arise from afferent synaptic connectivity contribute to balance performance and robustness.

Another direction for future research is to evaluate pruning in different types of topographic models \citep{poli2023introducing, lu2025end, zhang2025brain, deb2025toponets, blauch2022connectivity}. This can be done not only to achieve compression, but also to identify subnetworks that maintain high performance. Another possibility is to evaluate the existence of "winning tickets" as suggested by the lottery ticket hypothesis \citep{frankle2018lottery}, to test whether topographic networks are associated with sparser winning tickets. Extending topographic regularization to additional layers, including convolutional ones, could improve robustness to adversarial inputs, as suggested in \citet{bashivan2025learning}. In parallel, scaling topographic training to large-scale datasets could help facilitate comparisons of noise robustness between models and neural data \citep{jang2021noise, jang2024improved}.

Finally, we note that we studied topographic regularization in a supervised classification setting, where the spatial losses were jointly optimized with cross-entropy. More generally, topographic regularizers can be combined with many sorts of objectives, including unsupervised and self-supervised losses. An important direction for future research is to understand if the effects observed here generalize beyond supervised training.

\subsubsection*{Acknowledgments}
We thank Gerrit Sander for his comments on the manuscript.

\clearpage

\paragraph{Declaration of generative AI and AI-assisted technologies in the manuscript preparation process.}

During the preparation of this work the authors used Open AI's ChatGPT in order to assist with copyediting. After using this tool/service, the authors reviewed and edited the content as needed and takes full responsibility for the content of the published article.

\bibliography{iclr2025_conference}
\bibliographystyle{ieeetr}   % unsrt, plain, ieeetr, abbrv

\clearpage

\appendix
\setcounter{figure}{0}
\renewcommand{\thefigure}{A.\arabic{figure}}

\setcounter{table}{0}
\renewcommand{\thetable}{A.\arabic{table}}

\section{Appendix}
\subsection{Training Dynamics}

\paragraph{Accuracy.} 

\begin{figure}[H]
    \includegraphics[width=\linewidth]{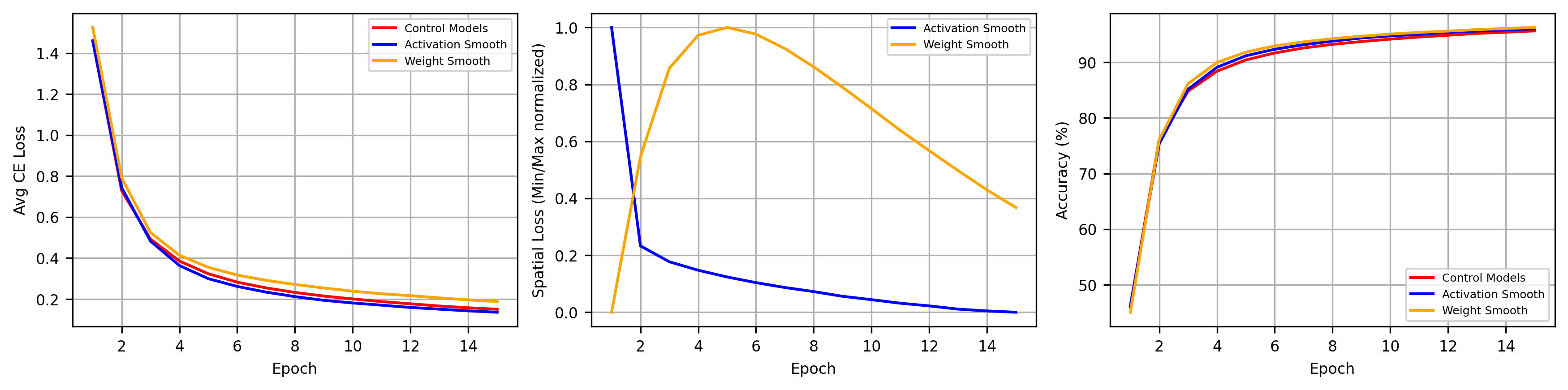}
    \caption{MNIST Train stats: Train-set accuracy and loss terms.}
    \label{fig:MNISTtrain}
\end{figure}

\begin{figure}[H]
    \includegraphics[width=\linewidth]{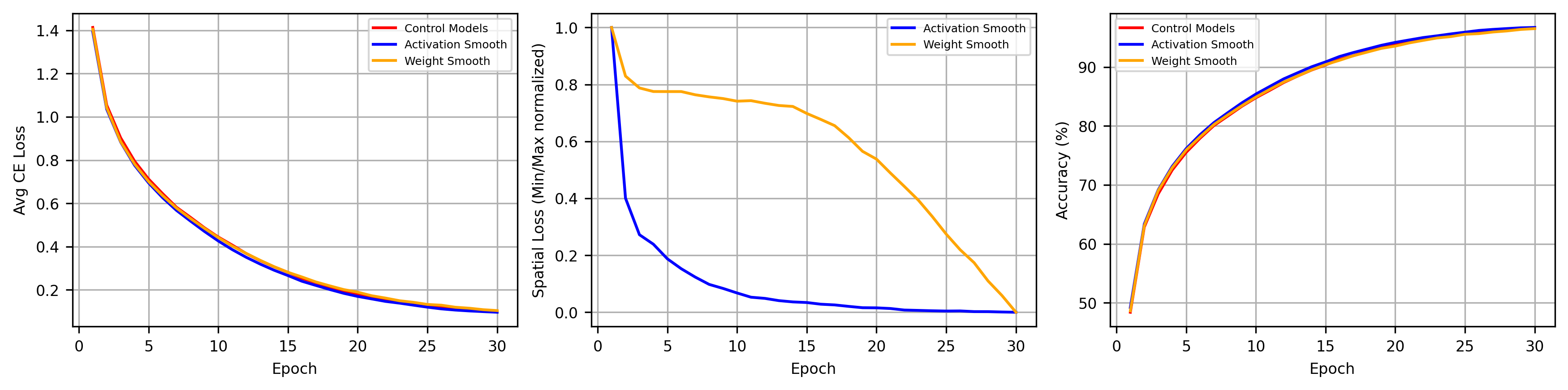}
    \caption{CIFAR-10 Train stats: Train-set accuracy and loss terms.}
    \label{fig:CIFARtrain}
\end{figure}

To evaluate how AS and WS regularization impacted training, we evaluated the cross-entropy loss and spatial-loss trajectory over the training epochs for $\lambda=0.1$. Figure \ref{fig:MNISTtrain} shows the results for MNIST, and Figure \ref{fig:CIFARtrain} shows similar findings for CIFAR-10.  For both MNIST and CIFAR-10, the trajectory of cross-entropy reduction (and accuracy) were highly similar, for all three types of models. This suggests that the different models learn the differentiation between classes at similar rates. With respect to the spatial loss terms, AS spatial-loss showed a strong early drop which rapidly asymptoted. For WS the dynamics were different: for MNIST, WS-loss showed an initial increase, followed by a decrease. For CIFAR-10 accuracy remained at a relatively high level (60\% of initial level) and began dropping when training accuracy was already high. These data suggest that the inclusion of AS and WS regularization did not strongly impact the dynamics of classification accuracy or those of cross entropy loss during training. They also suggest that WS loss functions can produce an initial trade-off between the spatial and cross-entropy objectives, perhaps because the topographic regularization harms the feature learning required for classification. Once the features are learned, the WS objective is more easily satisfied.

\newpage
\clearpage
\subsection{Global Similarity Comparison}
\label{sec:appendGlobalSim}
Let:
\begin{itemize}
  \item \( \mathbf{a}_i \in \mathbb{R}^B \) be the activation vector of unit \( i \) across a batch of size \( B \),
  \item \( S_{ij} = \cos(\mathbf{a}_i, \mathbf{a}_j) = \dfrac{\mathbf{a}_i \cdot \mathbf{a}_j}{\|\mathbf{a}_i\| \, \|\mathbf{a}_j\|} \) be the cosine similarity between units \( i \) and \( j \),
  \item \( d_{ij} \) :  the Euclidean distance between units \( i \) and \( j \) on a fixed \( 11 \times 11 \) spatial grid (so \( N = 121 \) units),
  \item \( A_{ij} = \dfrac{1}{d_{ij} + 1} \) : the target similarity between two units based on spatial proximity.
\end{itemize}

The spatial loss, as implemented by  \citet{poli2023introducing} is defined as:

\[
\mathcal{L}_{\text{spatial}} = \frac{1}{N(N - 1)} \sum_{\substack{i, j = 1 \\ i \ne j}}^N \left( S_{ij} - A_{ij} \right)^2 = \frac{1}{N(N - 1)} \sum_{\substack{i, j = 1 \\ i \ne j}}^N \left( S_{ij} - \frac{1}{d_{ij} + 1} \right)^2
\]

\newpage
\clearpage

\begin{figure}[H]
    \centering
    \includegraphics[width=.7\linewidth]{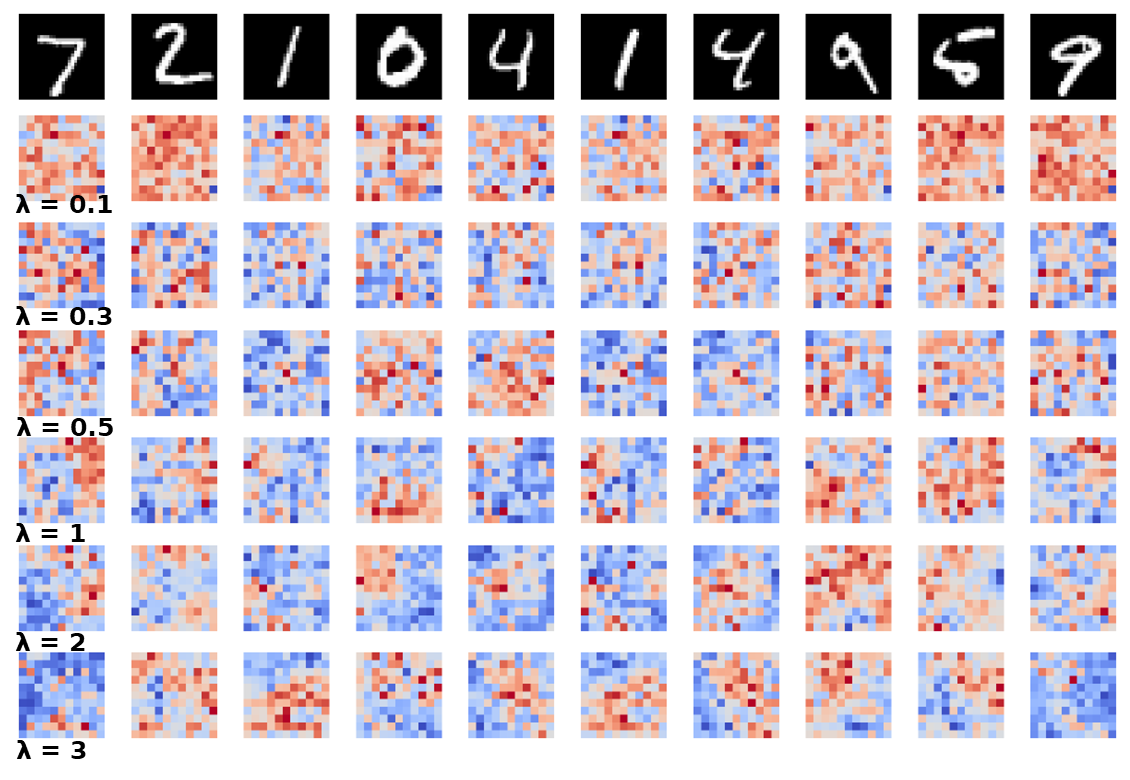}
    \caption{\textbf{MNIST topography examples under global activation similarity regularization}.  Example activation maps from the topographic grid layer are shown for MNIST inputs across increasing values of the regularization parameter $\lambda$ for models trained with a global activation similarity loss. Columns: different input image; Rows: different values of $\lambda$.}
    \label{fig:globalsimMNISTtopo}
\end{figure}

\begin{figure}[H]
    \centering
    \includegraphics[width=0.7\linewidth]{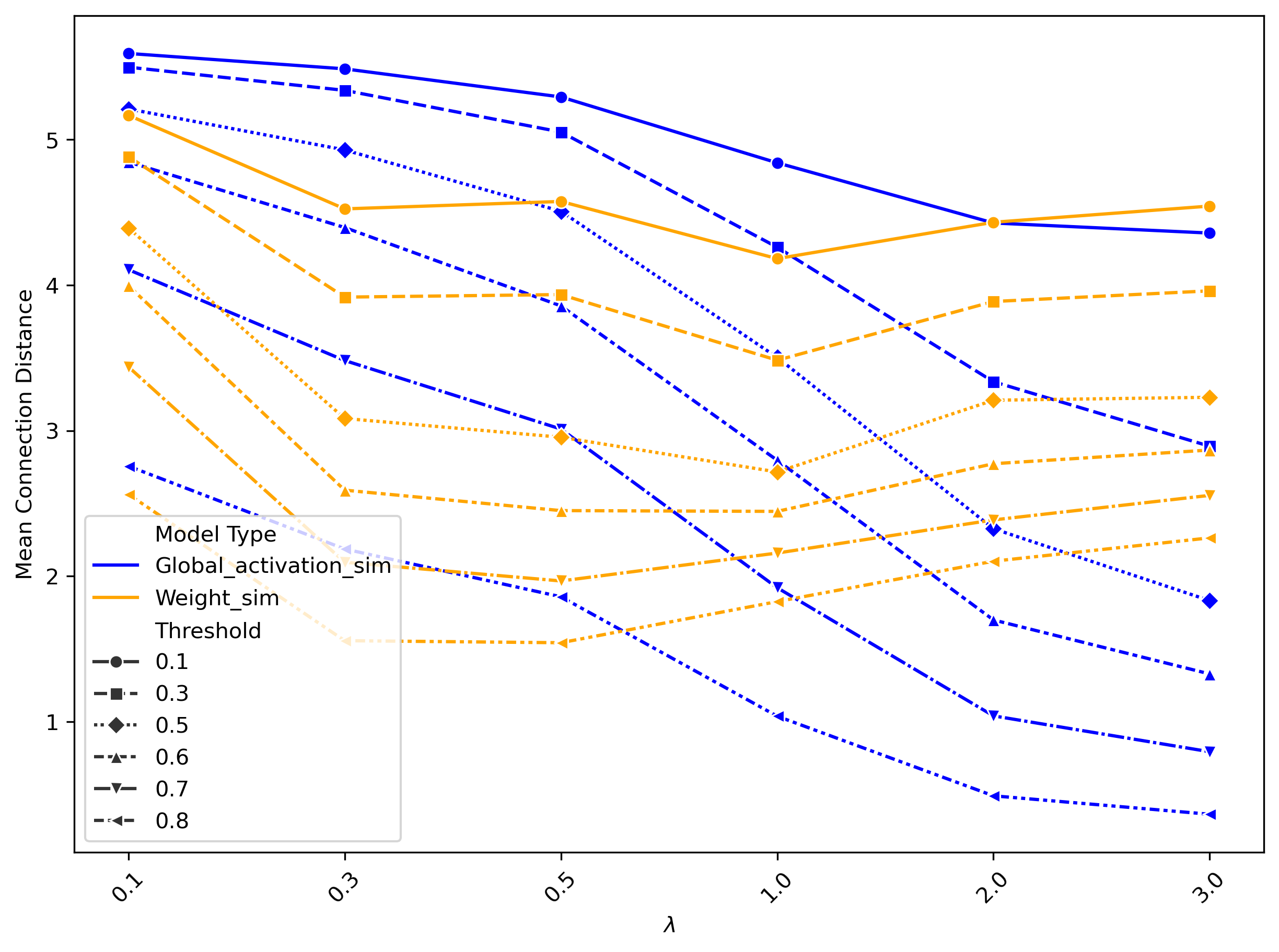}
    \caption{\textbf{Colocalization under global similarity regularization}.  Mean spatial distance between pairs of units whose activity correlation exceeds a threshold $\alpha \in \{0.1, 0.3, 0.5, 0.6, 0.7, 0.8\}$ as a function of $\lambda$. Distances are reported for models trained with global activation similarity (blue) and weight similarity (yellow) losses. Curves correspond to distances computed at individual correlation thresholds $\alpha$.}
    \label{fig:globalsimcolo}
\end{figure}
\newpage
\clearpage

\subsection{Supplementary Figures}

\begin{figure}[H]
  \centering
  \begin{tabular}{cc}
    \includegraphics[width=0.49\textwidth]{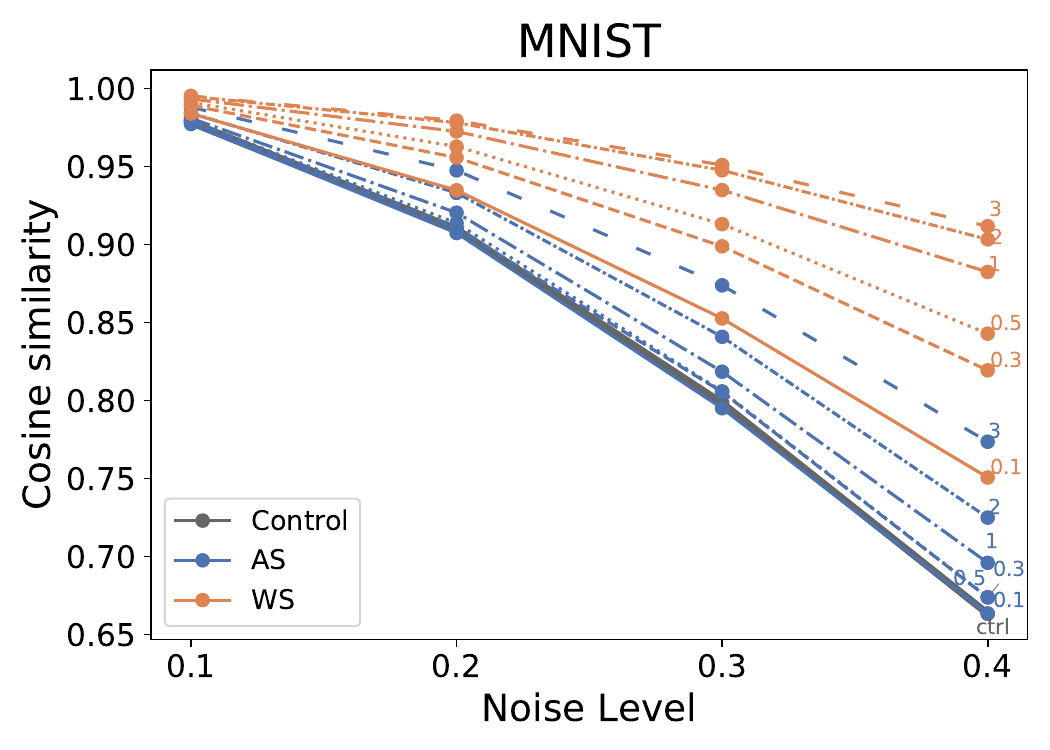} & \includegraphics[width=0.49\textwidth]{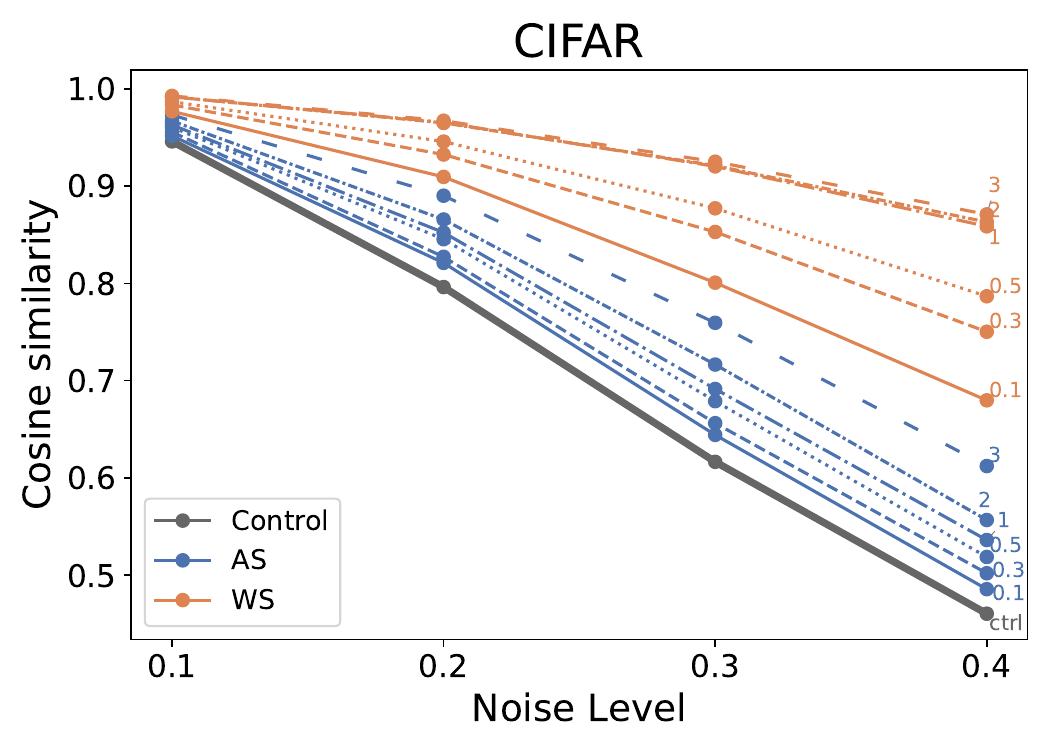} \\ 
    \includegraphics[width=0.49\textwidth]{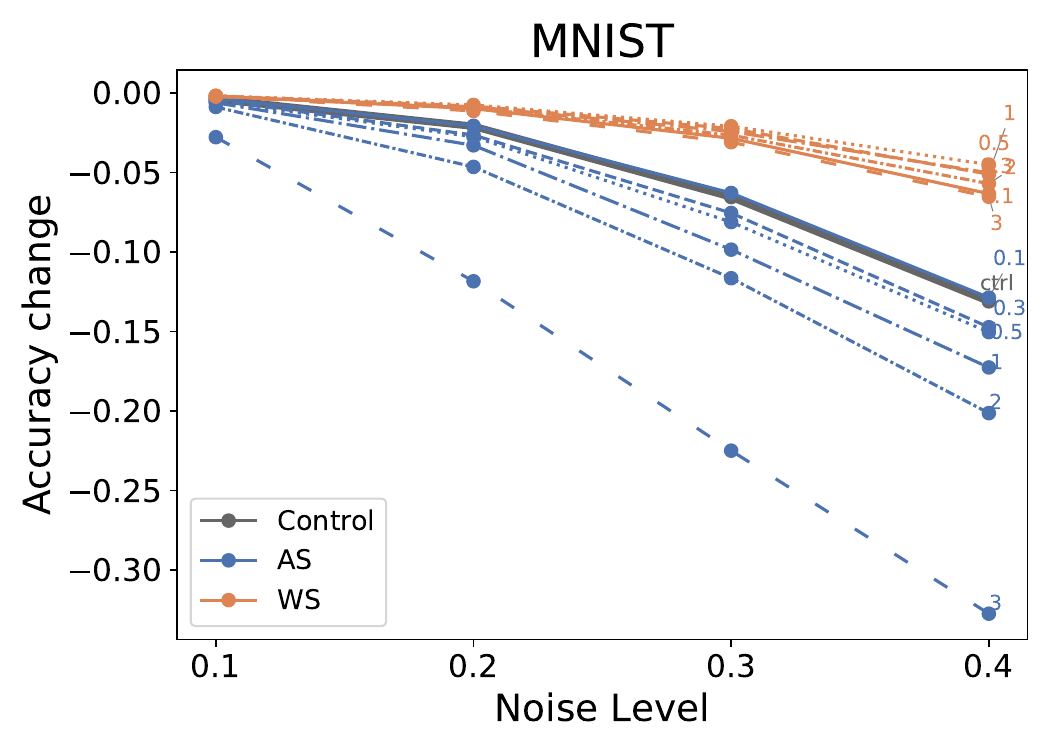} & \includegraphics[width=0.49\textwidth]{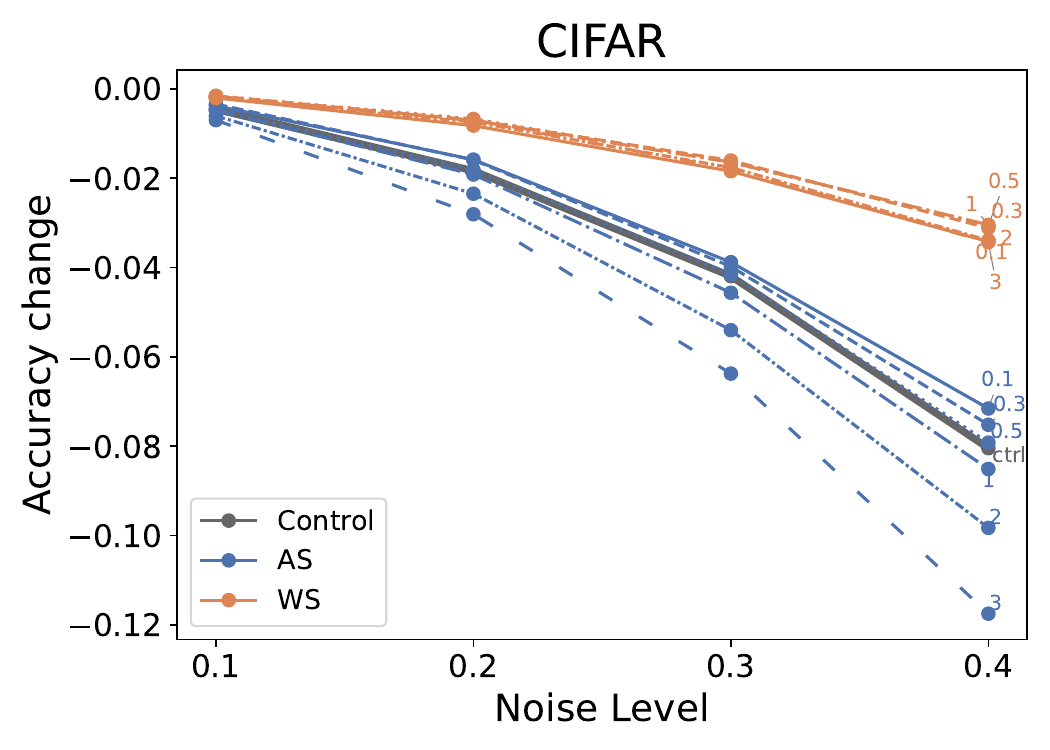} \\ 
  \end{tabular}
  \caption{\textbf{Robustness to weight perturbations evaluated across individual values of $\lambda$}. 
Robustness is assessed by changes in representational geometry (top row) and test accuracy (bottom row) under increasing levels of additive noise applied to the final classification-layer weights. Representational geometry is defined as the $10 \times 10$ cosine-similarity matrix computed from category-level weight vectors, and robustness is quantified as the similarity between the original and perturbed matrices. Results are shown separately for MNIST (left) and CIFAR (right), and for control, AS, and WS models at individual values of the regularization parameter $\lambda$ (indicated by line style).}
    \label{fig:Approbweight}
\end{figure}

\begin{figure}[H]
  \centering
  \begin{subfigure}{0.49\textwidth}
    \includegraphics[width=\linewidth]{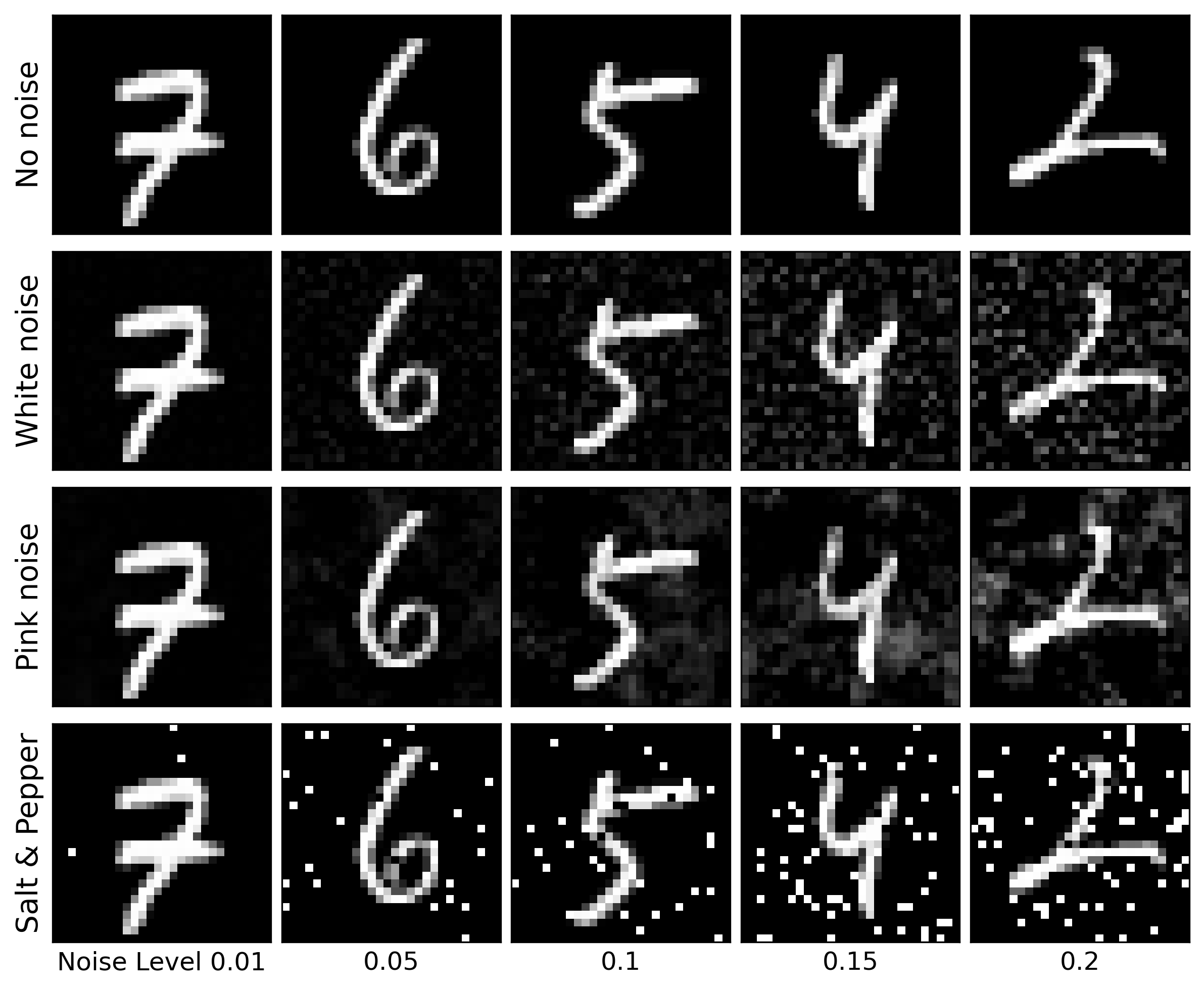}
    \caption{MNIST}
    \label{fig:noise_example_mnist}
  \end{subfigure}
  \hfill
  \begin{subfigure}{0.49\textwidth}
    \includegraphics[width=\linewidth]{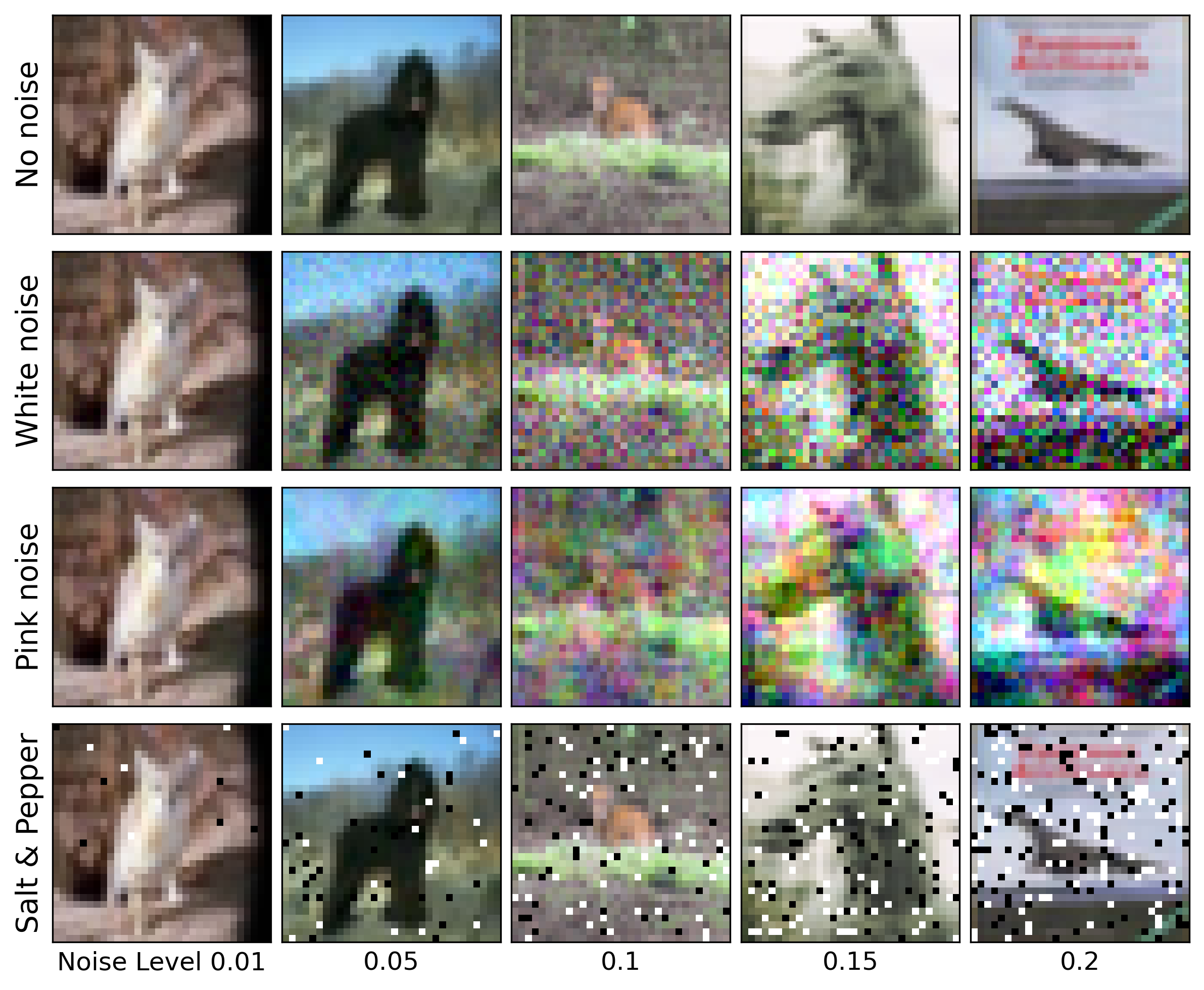}
    \caption{CIFAR-10}
    \label{fig:noise_example_cifar}
  \end{subfigure}
  \caption{\textbf{Examples of noise types and noise levels}. Representative MNIST (a) and CIFAR-10 (b) images are shown under different noise types (white noise, pink noise, and salt-and-pepper noise) and increasing noise levels. Rows: noise types; columns: noise level. For reference, noise-free inputs are presented in the top row.}
  \label{fig:noise_example}
\end{figure}

\begin{figure}[H]
  \centering
  \begin{subfigure}[b]{\textwidth}
    \centering
    \includegraphics[width=\textwidth]{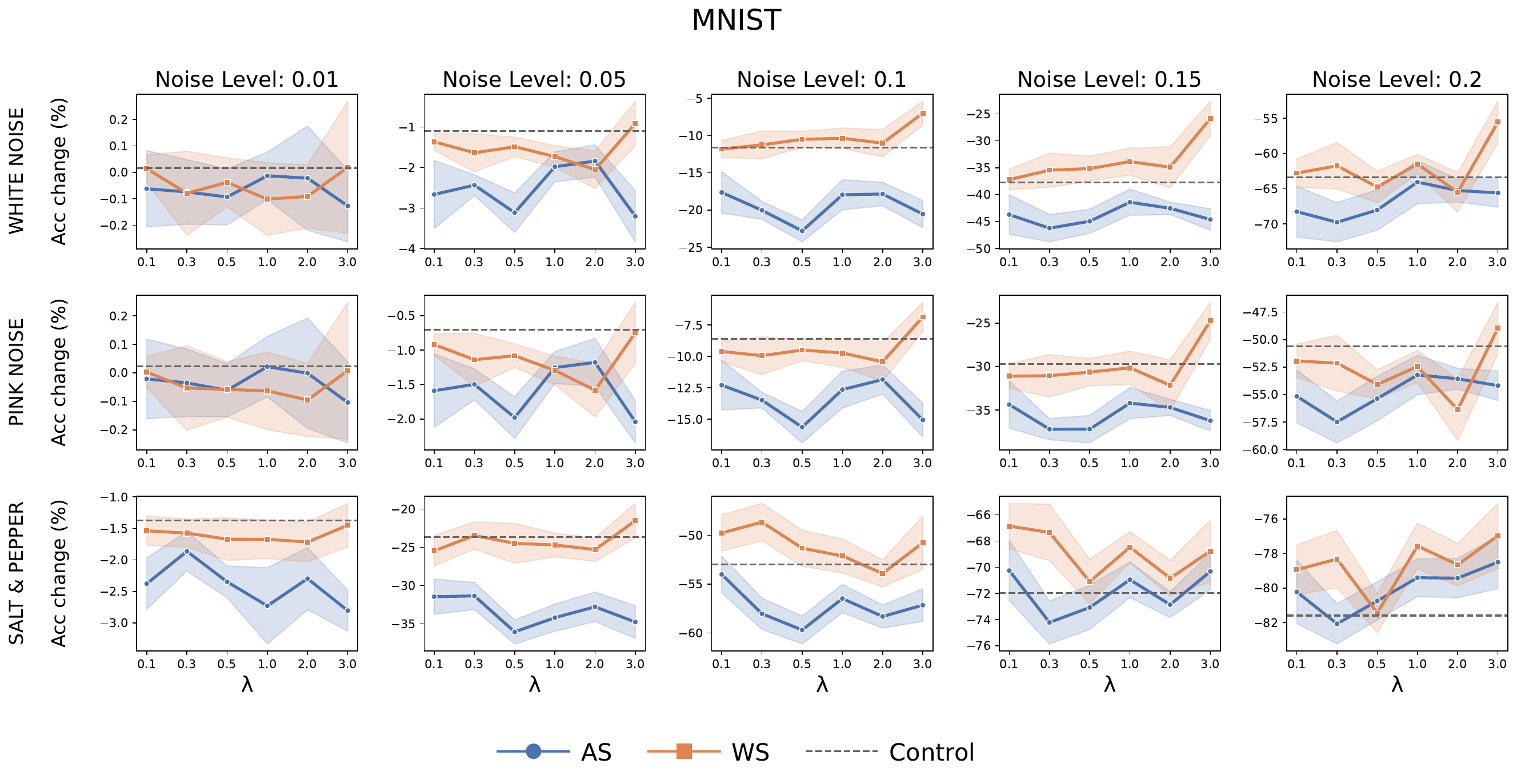}
    \label{fig:noise_acc_mnist}
  \end{subfigure}
  \vspace{0.5em}
  \begin{subfigure}[b]{\textwidth}
    \centering
    \includegraphics[width=\textwidth]{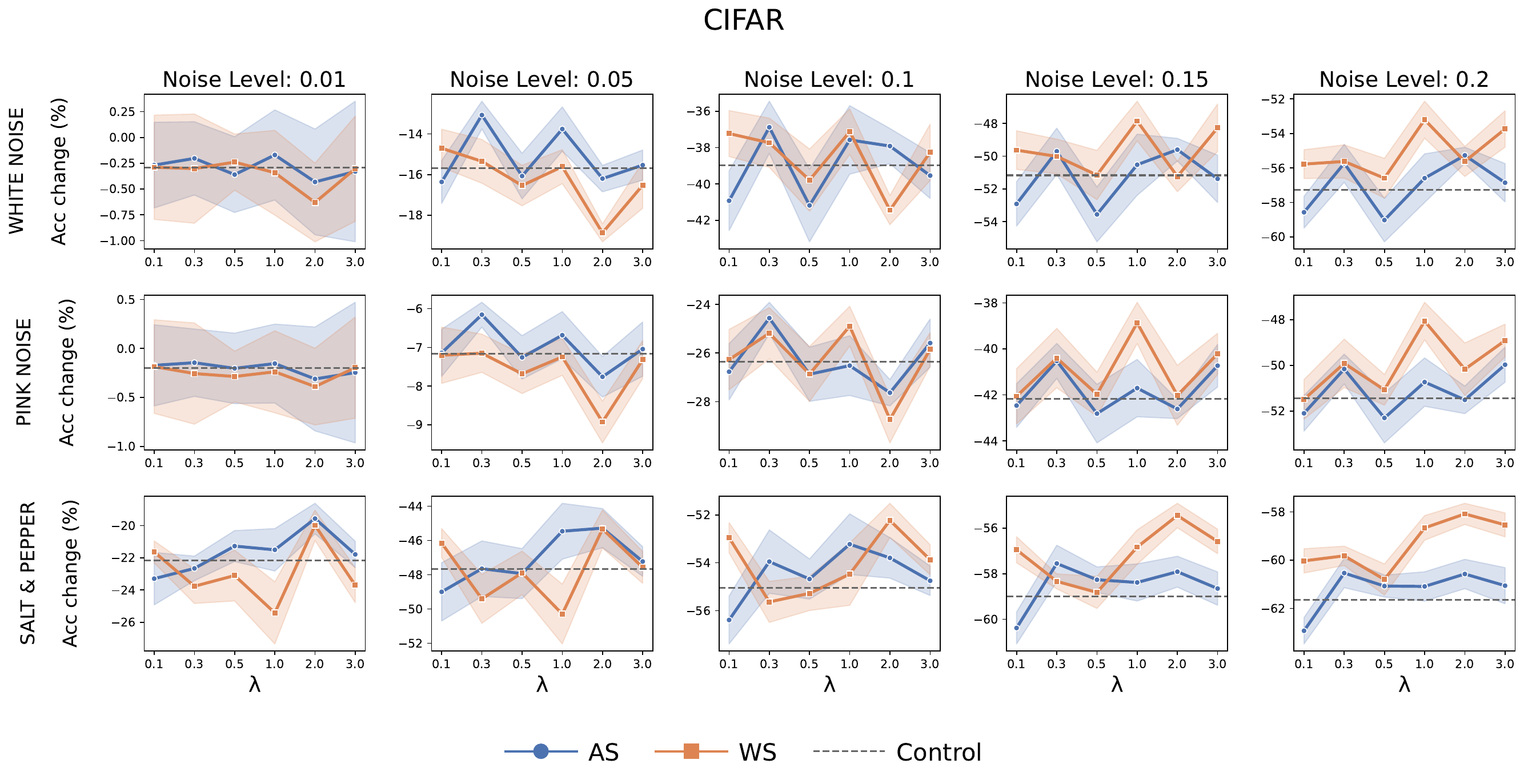}
    \label{fig:noise_acc_cifar}
  \end{subfigure}
  \caption{\textbf{Accuracy changes under input corruption across noise type, noise level, and topographic regularization}. 
Changes in classification accuracy relative to baseline (non-perturbed) models are shown as a function of $\lambda$ for models trained with AS, WS, and control objectives. Results are shown for MNIST (top) and CIFAR (bottom), for three types of corruption (white noise, pink noise, and salt-and-pepper; rows), and for increasing noise levels (columns). Accuracy values are normalized relative to the corresponding noise-free baseline. Shaded regions indicate $\pm s.e.m$ across runs.}
  \label{fig:Appnoise_acc}
\end{figure}

\begin{figure}[H]
    \begin{subfigure}[t]{0.48\textwidth}
        \includegraphics[width=\textwidth]{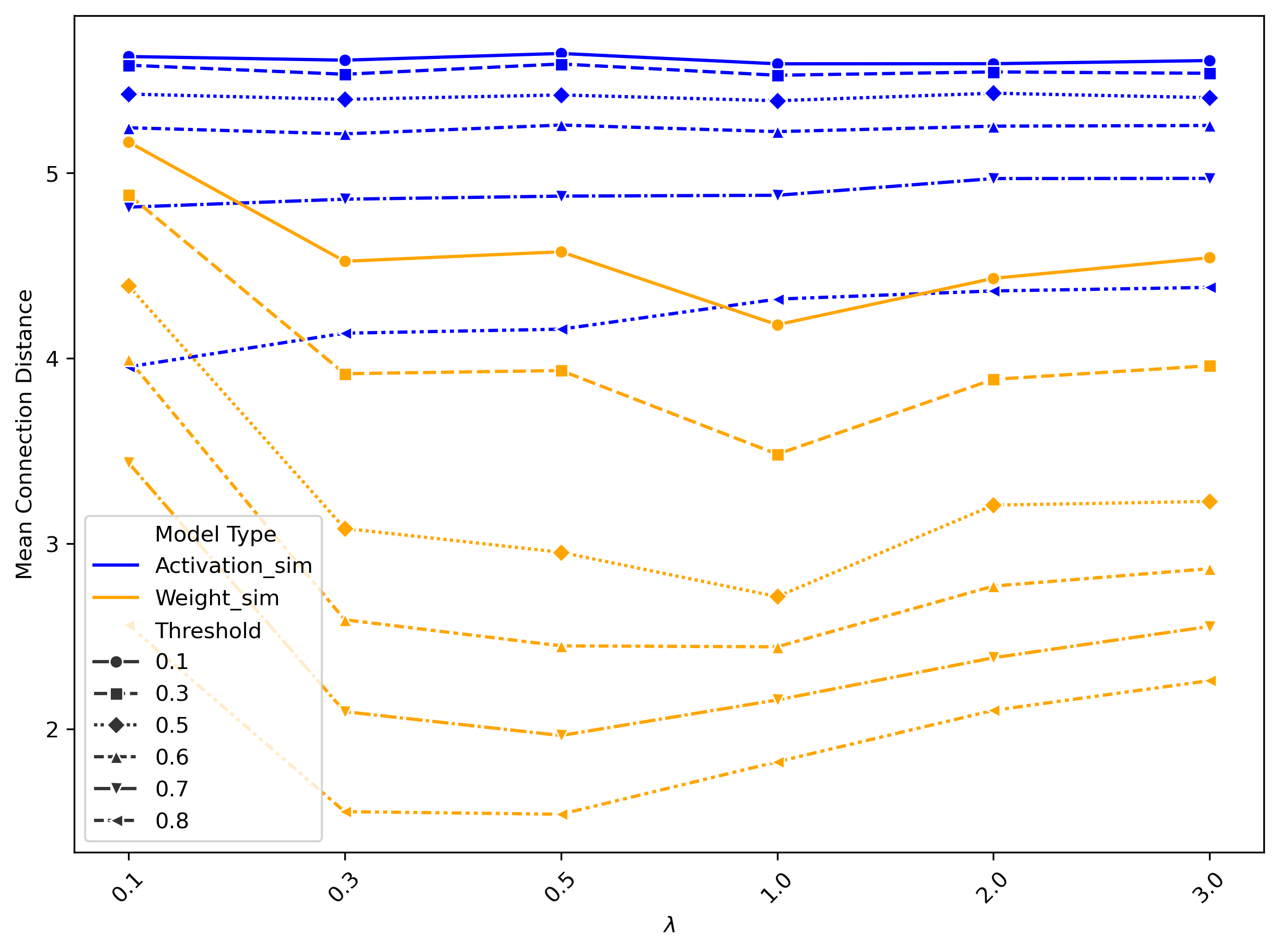}
        % \caption{MNIST}
        \label{fig:mnist_connect_dist}
    \end{subfigure}
    \hfill
    \begin{subfigure}[t]{0.48\textwidth}
        \includegraphics[width=\textwidth]{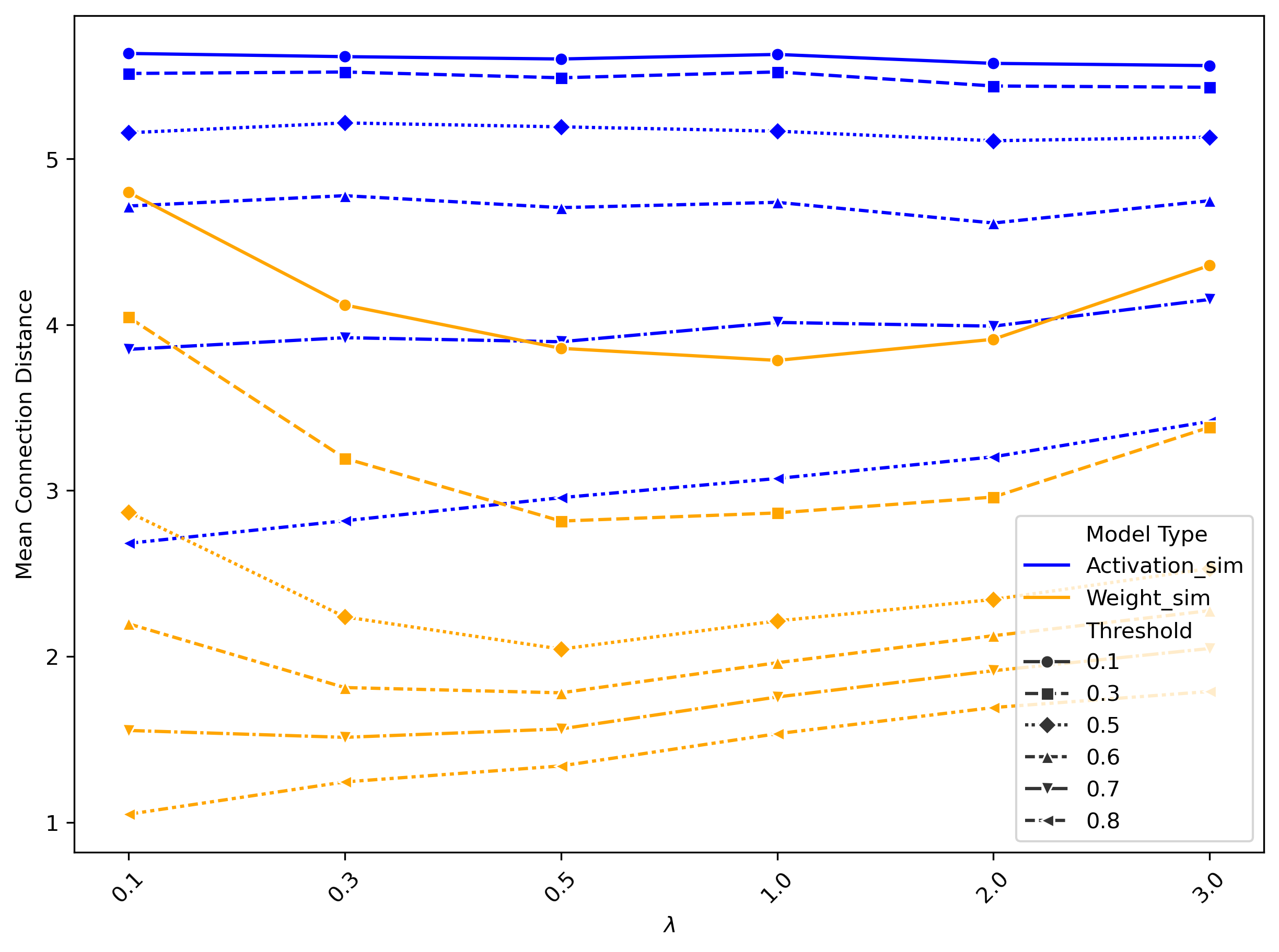}
        % \caption{CIFAR-10}
        \label{fig:cifar_connect_dist}
    \end{subfigure}
    \caption{\textbf{Correlated-unit distances across individual correlation thresholds}.  Mean spatial distance between pairs of units whose activity correlation exceeds a threshold $\alpha \in \{0.1, 0.3, 0.5, 0.6, 0.7, 0.8\}$ is shown for MNIST (left) and CIFAR (right). Separate curves corresponding to individual correlation thresholds.}
    \label{fig:Appconnect_dist}
\end{figure}

\begin{figure}[H]
    \centering
    % ---------- First image (2/3 page width) ----------
    \begin{minipage}[t]{0.60\textwidth}
        \centering
        \includegraphics[width=\linewidth]{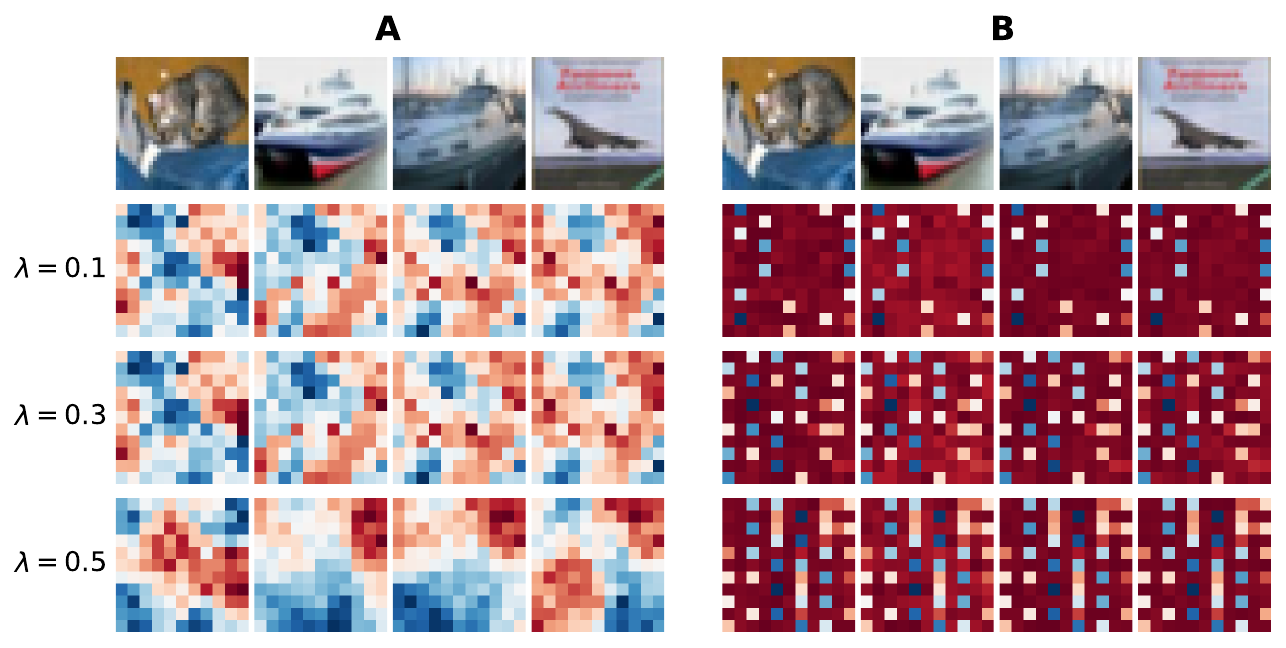}
        % \captionof{subfigure}{Main comparison}
    \end{minipage}
    \hfill
    % ---------- Second image (1/3 page width) ----------
    \begin{minipage}[t]{0.35\textwidth}
        \centering
        \includegraphics[width=\linewidth]{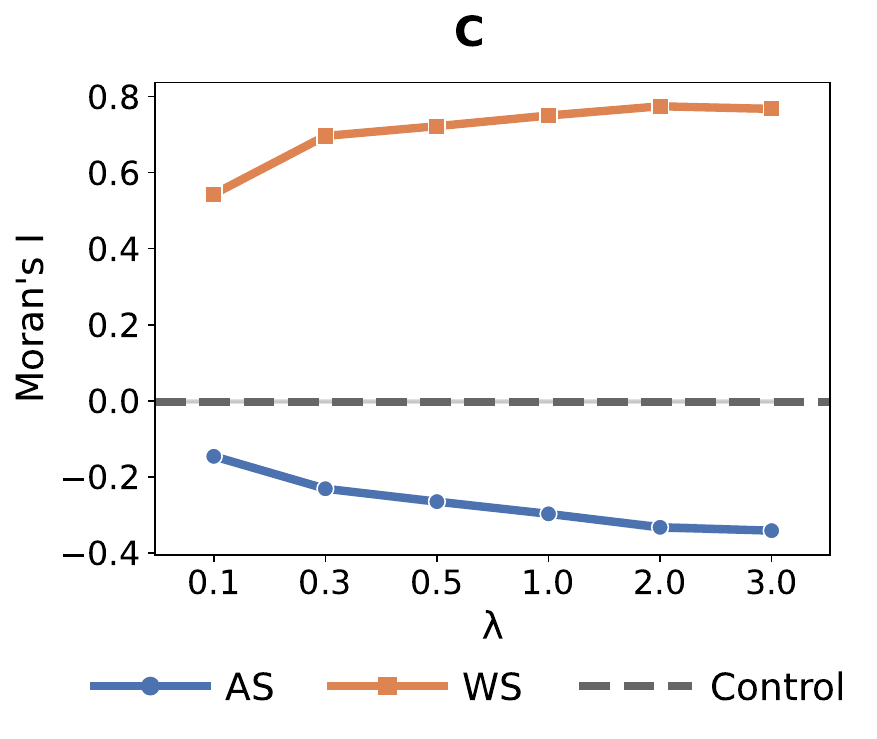}
        % \captionof{subfigure}{Auxiliary result}
    \end{minipage}

    \caption{\textbf{Spatial smoothness of activation maps under activation- and weight-similarity training for CIFAR-10}.}
    \label{fig:Appactivation_comparison_cifar}
\end{figure}

\begin{figure}[H]
    \includegraphics[width=\linewidth]{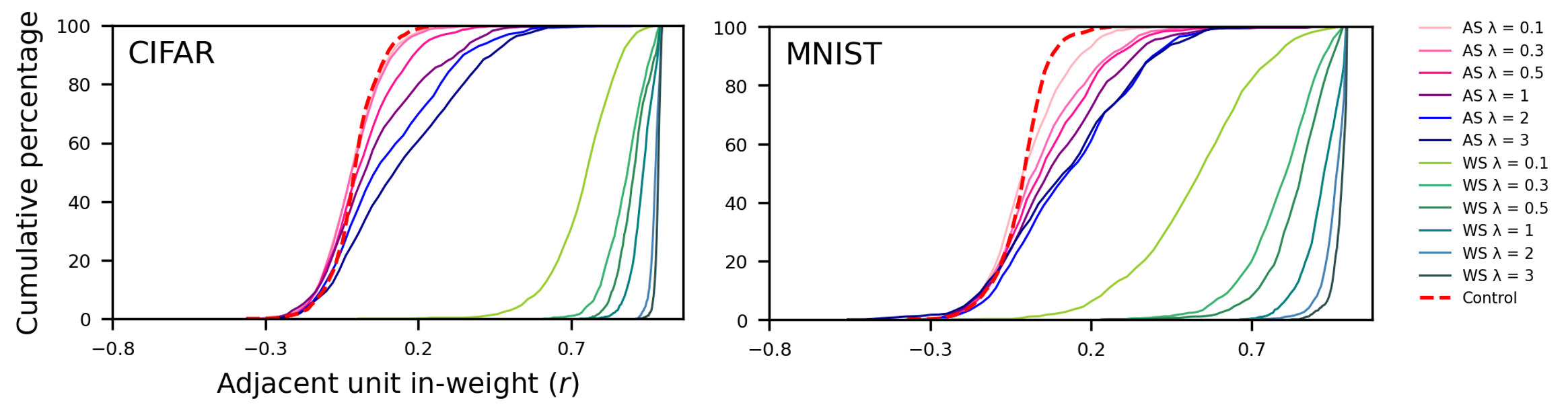}
    \caption{\textbf{Incoming weight correlations between adjacent units}.  Cumulative distributions of average pairwise correlations between incoming weight vectors of adjacent units. For each unit, the mean correlation is computed form all adjacent neighbors. Distributions are shown for control, AS, and WS models across values of the regularization parameter $\lambda$.}
    \label{fig:bothmodels_incomingWeights}
\end{figure}

\begin{figure}[H]
    \centering
    \includegraphics[width=1\linewidth]{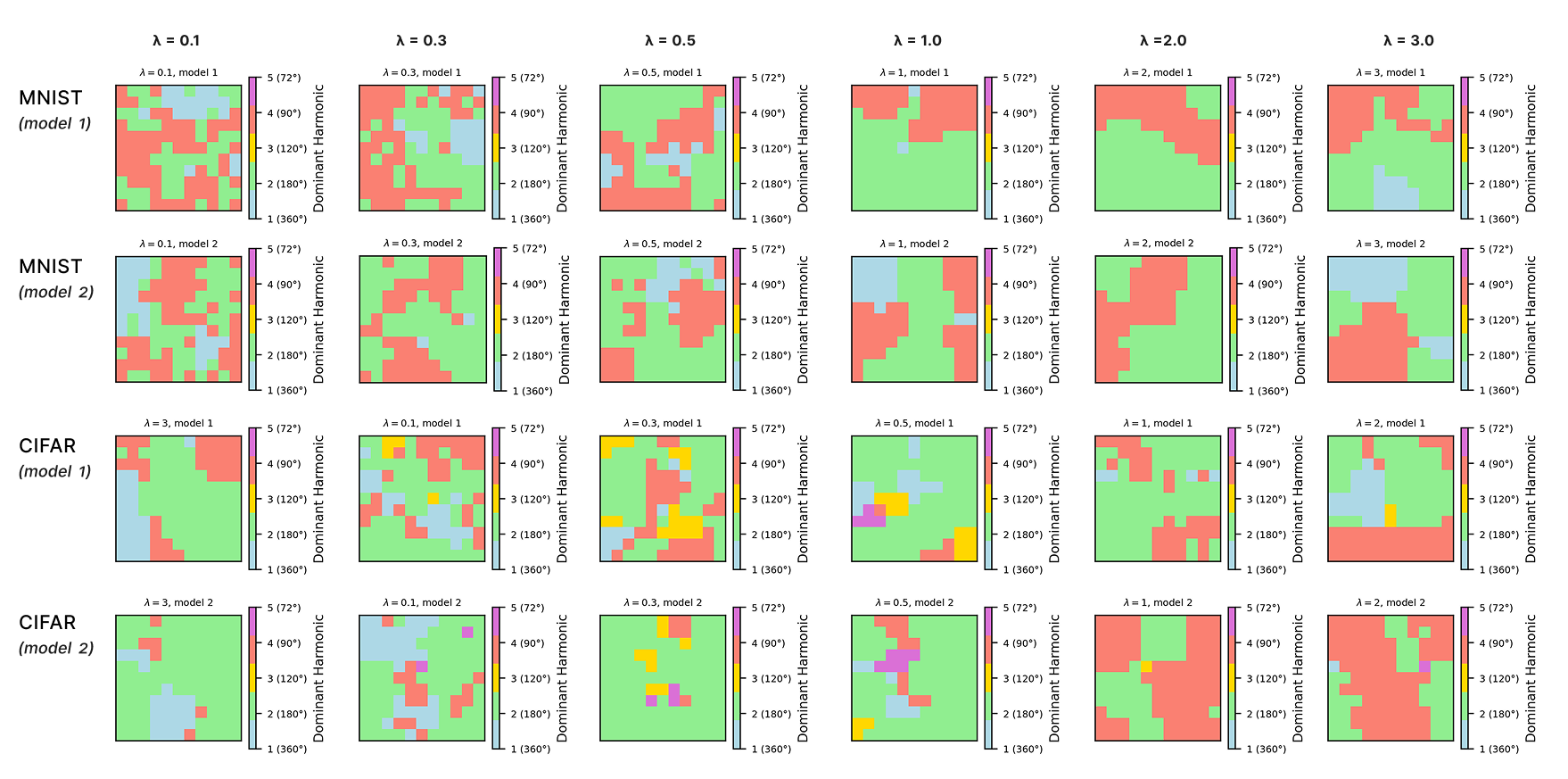}
    \caption{\textbf{Angular response properties under weight-similarity training}. Topographic maps show the dominant angular response type for units in the grid across values of the regularization parameter $\lambda$ under WS training. Angular response types correspond to five harmonic components with periodicities of 360°, 180°, 120°, 90°, and 72° (Cycles 1–5), indicated by color. Results are shown for MNIST and CIFAR. Two randomly trained models shown per dataset. Columns correspond to values of $\lambda$.}
    \label{fig:AppWedgeHarmonics}
\end{figure}

\begin{figure}[H]
    \centering
    \includegraphics[width=1\linewidth]{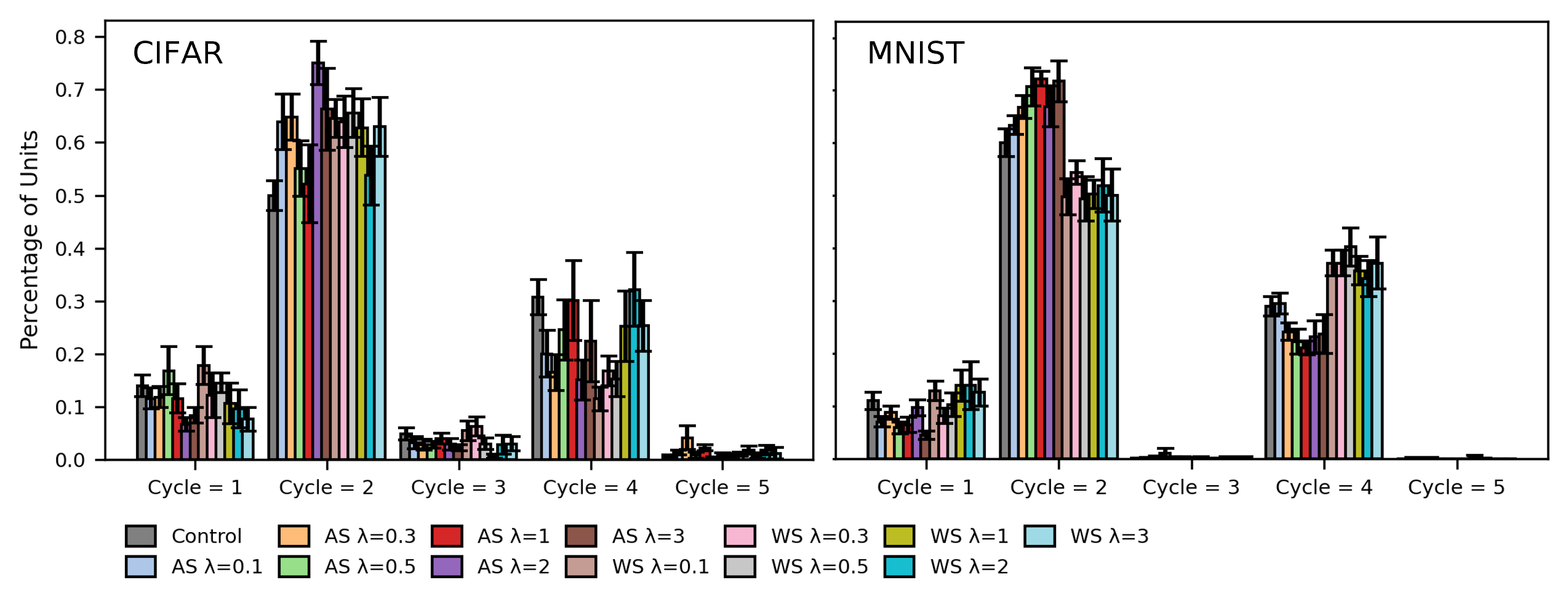}
    \caption{\textbf{Angular tuning profiles across training conditions and regularization strengths ($\lambda$}.   Percentage of units assigned to each angular tracking profile (Cycles 1–5) is shown. Cycles correspond to harmonic response types with periodicities of 360°, 180°, 120°, 90°, and 72°, respectively. Within each cycle, bar order is control, followed AS and WS, at increasing values of $\lambda$. Bars indicate mean percentages across runs; error bars denote variability across runs.}
    \label{fig:AppharmonicsCIFARMNIST}
\end{figure}

\begin{comment}
    \begin{figure}[H]
    \centering
    \includegraphics[width=1\linewidth]{imgs/actmaps/cycle4tuning.png}
    \caption{\texttt{cycle=4} tuning profiles.}
    \label{fig:appendix_cycle4}
\end{figure}

\end{comment}

\begin{figure}[H]
    \centering
    \includegraphics[width=1\linewidth]{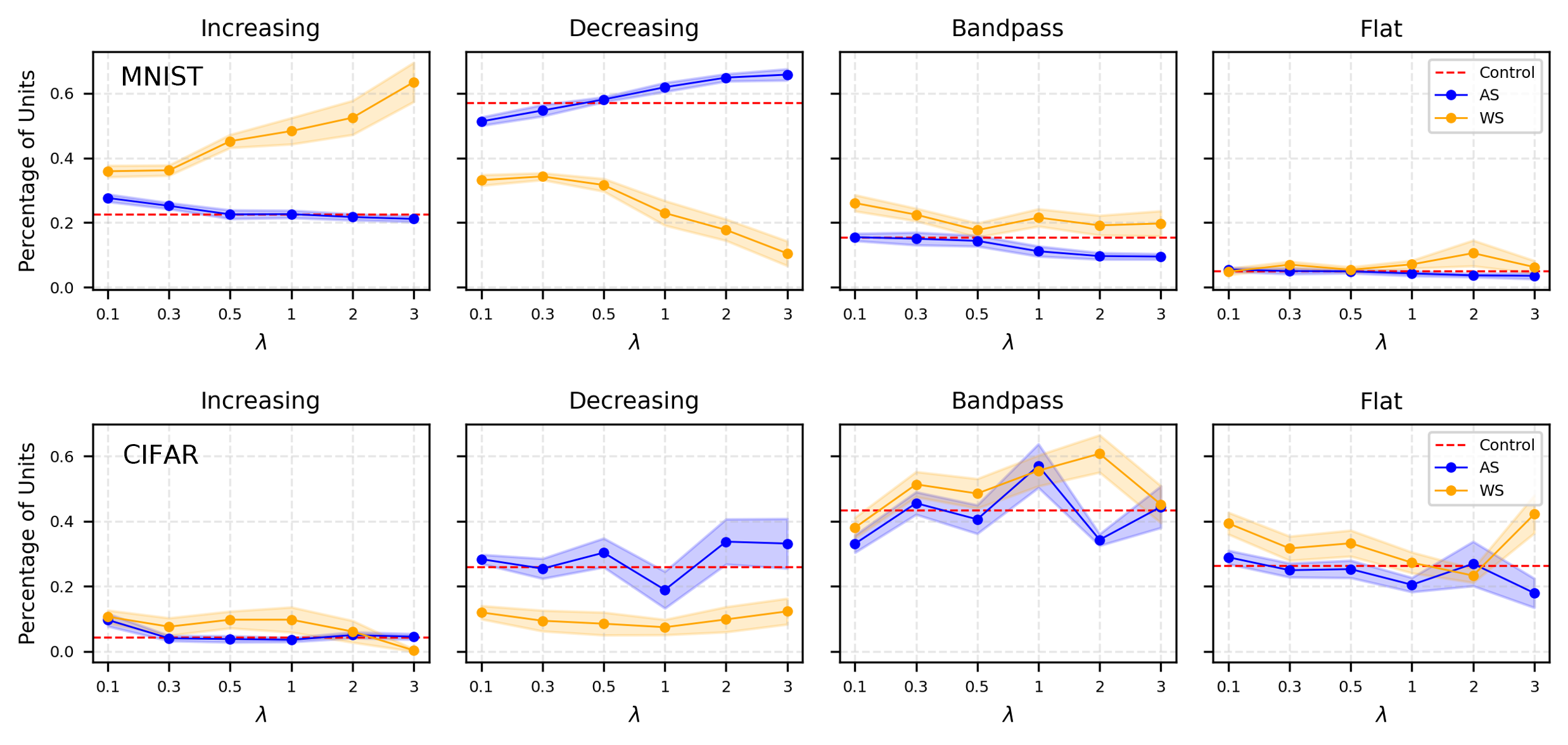}
    \caption{\textbf{Eccentricity tuning profiles across training conditions and regularization strengths}. Percentage of units showing increasing, decreasing, band-pass, or flat eccentricity tuning profiles. Each panel corresponds to one eccentricity tuning profile, as defined in the main text. \textit{Increasing} and \textit{decreasing} profiles correspond to monotonic changes in response magnitude with eccentricity and reflect units with positive or negative radial gain, respectively.}
    \label{fig:Appeccentricity_allprofiles}
\end{figure}

\begin{comment}
    \begin{figure}[H]
    \centering
    \includegraphics[width=1\linewidth]{imgs/actmaps/bandpass_by_eccentricity_CIFAR.png}
    \caption{Preference for eccentricity in units showing bandpass responses for CIFAR-10 training.}
    \label{fig:Appendix_allcccalib}
\end{figure}

\end{comment}

\begin{figure}[H]
    \centering
    \includegraphics[width=\linewidth]{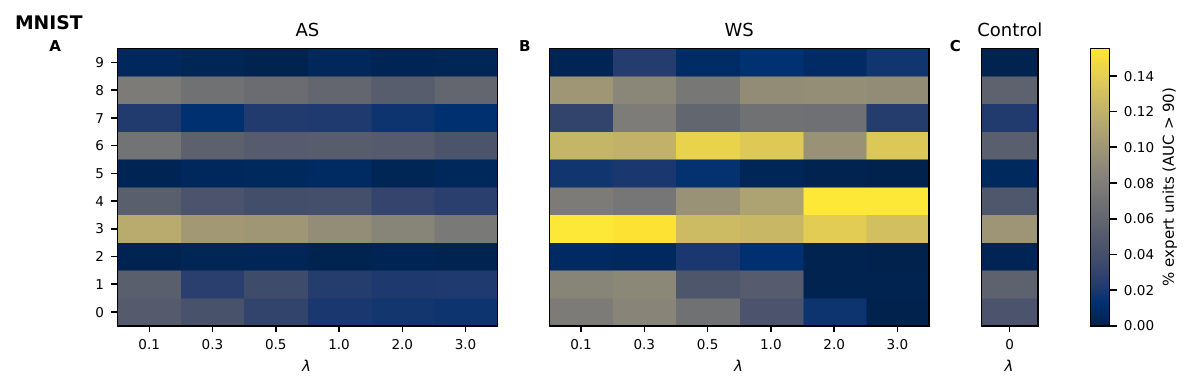}
    \includegraphics[width=\linewidth]{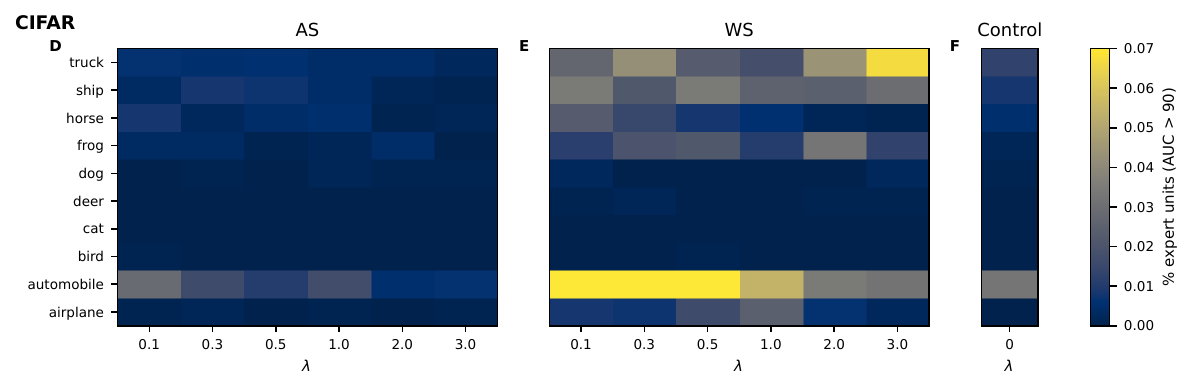}
  \caption{\textbf{Category-wise distribution of high-expertise units across training conditions}. 
Heatmaps show the percentage of units with high category discriminability (AUC $>0.90$) for each category as a function of the regularization parameter $\lambda$. For AS and WS models, columns correspond to values of $\lambda$; control models are shown as $\lambda = 0$. Color scale indicates the proportion of high-expertise units within each category.}
    \label{fig:app_expert_distrib}
\end{figure}

\begin{comment}
    \begin{figure}[H]
    \centering
    % First subfigure
    \begin{subfigure}{\linewidth}
        \centering
        \includegraphics[width=\linewidth]{imgs/compProp/mnist_grouped_condition_similarity.png}
    \end{subfigure}
    
    \vspace{0.5em} % small vertical space between subfigures
    
    % Second subfigure
    \begin{subfigure}{\linewidth}
        \centering
        \includegraphics[width=\linewidth]{imgs/compProp/cifar_grouped_condition_similarity.png}
    \end{subfigure}
    
    \caption{Comparison of grouped condition similarity on (a) MNIST and (b) CIFAR datasets.}
    \label{fig:Appendix_overallRSA}
\end{figure}

\end{comment}

\end{document}